%% file: ms.tex
\pdfoutput=1

\documentclass[a4paper, 11pt, abstracton, titlepage, oneside]{scrartcl}
\makeatletter

\input{resources/packages.tex}
\theoremstyle{plain} \newtheorem{result}{Result}
\input{resources/configuration.tex}
\pgfplotsset{compat=1.18}
\begin{document}
\emergencystretch 3em
\input{./resources/abbreviations}
\input{./resources/notation}
\input{./resources/titlepage.tex}
\input{./contents/introduction}
\input{./contents/relatedwork}
\input{./contents/problemsetting}
\input{./contents/methodology}
\input{./contents/casestudy}
\input{./contents/results}
\input{./contents/conclusion}


%
\singlespacing{
\bibliographystyle{model5-names}
\bibliography{ms}} 
\newpage

\appendix
\renewcommand{\thesection}{\Alph{section}}
\renewcommand{\thesubsection}{\Alph{section}.\arabic{subsection}}

\onehalfspacing
	\normalsize
        \input{./contents/appendix}
\end{document}

%% file: resources/packages.tex
\usepackage[T1]{fontenc}
\usepackage[utf8]{inputenc}
\usepackage[onehalfspacing]{setspace}
\usepackage[headsepline]{scrlayer-scrpage}
\clearpairofpagestyles 
\usepackage{layout}
\usepackage{geometry}
\usepackage{adjustbox}
\usepackage{authblk}
\usepackage[titletoc,title]{appendix}

\usepackage[hyphens]{url}
\usepackage{color}

\usepackage{amsmath,amssymb,amsfonts,amsthm, mathtools, bm}
\usepackage{thmtools, thm-restate}
\usepackage{nicefrac}
\usepackage{array}

\usepackage[normal]{threeparttable}
\usepackage{threeparttablex}
\usepackage{tabularx}
\usepackage{longtable}
\usepackage{booktabs}
\usepackage{colortbl}
\usepackage{multirow}
\usepackage{makecell}

\usepackage[acronym]{glossaries}

\usepackage{subcaption}
\usepackage{floatrow}
\usepackage{graphicx}
\usepackage{placeins}

\usepackage{tikz}
\usetikzlibrary{arrows.meta}
\usetikzlibrary{math}
\usetikzlibrary{shapes}
\usepackage{pgfplots}
\usepgfplotslibrary{groupplots}
\usepgfplotslibrary{dateplot}
\usepackage{tikzscale}

\usepackage{enumerate}
\usepackage{multicol}
\usepackage[author=Patrick]{pdfcomment}
\usepackage{xparse}
\usepackage{twoopt}
\usepackage{etoolbox}

\usepackage{eurosym}
\usepackage{ellipsis}
\usepackage{natbib}
\usepackage{paralist}
\usepackage{ulem}

\usepackage{nameref}
\usepackage{amsfonts}
\usepackage{amsmath}
\usepackage{adjustbox}
\usepackage{algorithm}
\usepackage{algpseudocode}
\usepackage{multirow}
\usepackage{xurl} 
\usepackage{bm}
\usepackage{todonotes}
\usepackage{setspace}
\usepackage{amsthm}

%% file: resources/configuration.tex
\AtBeginDocument{
	\newgeometry{
		textheight=9in,
		textwidth=6.5in,
		top=1in,
		headheight=14pt,
		headsep=25pt,
		footskip=30pt
	}
}
\newsavebox{\abstractbox}
\renewenvironment{abstract}
{\begin{lrbox}{0}\begin{minipage}{\textwidth}
			\begin{center}\normalfont\sectfont\abstractname\end{center}\quotation}
		{\endquotation\end{minipage}\end{lrbox}%
	\global\setbox\abstractbox=\box0 }

\makeatletter
\expandafter\patchcmd\csname\string\maketitle\endcsname
{\vskip\z@\@plus3fill}
{\vskip\z@\@plus2fill\box\abstractbox\vskip\z@\@plus1fill}
{}{}
\makeatother

\addtokomafont{captionlabel}{\footnotesize\bfseries} 
\addtokomafont{caption}{\footnotesize\bfseries} 

\bibpunct[, ]{(}{)}{,}{a}{}{,}%
\newtheorem{definition}{Definition}[section]

\let\theoremstyle\relax

\counterwithin*{equation}{section}

\newenvironment{myfont}{\fontfamily{ccr}\selectfont}{\par}
\DeclareTextFontCommand{\textmyfont}{\myfont}

\AtBeginDocument{%
	\abovedisplayskip=7.5pt plus 0pt minus 0pt
	\abovedisplayshortskip=7.5pt plus 0pt minus 0pt
	\belowdisplayskip=7.5pt plus 0pt minus 0pt
	\belowdisplayshortskip=7.5pt plus 0pt minus 0pt
}

\newcolumntype{L}[1]{>{\raggedright\let\newline\\\arraybackslash\hspace{0pt}}p{#1}}
\newcolumntype{C}[1]{>{\centering\let\newline\\\arraybackslash\hspace{0pt}}p{#1}}
\newcolumntype{R}[1]{>{\raggedleft\let\newline\\\arraybackslash\hspace{0pt}}p{#1}}
\def\checkmark{\tikz\fill[scale=0.4](0,.35) -- (.25,0) -- (1,.7) -- (.25,.15) -- cycle;}

\renewcommand{\emph}[1]{\textit{#1}}

%% file: resources/titlepage.tex
\title{\large Ambulance Demand Prediction via Convolutional Neural Networks}

\author[1]{\normalsize Maximiliane Rautenstrauß}
\author[2]{\normalsize Maximilian Schiffer}
\affil{\small 
	School of Management, Technical University of Munich
	
	\scriptsize maximiliane.rautenstrauss@tum.de
	
	\small
	\textsuperscript{2}School of Management \& Munich Data Science Institute, Technical University of Munich
	
	\scriptsize schiffer@tum.de}

\date{}

\lehead{\pagemark}
\rohead{\pagemark}

\begin{abstract}
\begin{singlespace}
{\small\noindent \input{./contents/abstract.tex}\\ \smallskip}
{\footnotesize\noindent \textbf{Keywords:} AI in health care, spatio-temporal forecasting, convolutional neural networks}
\end{singlespace}
\end{abstract}

\maketitle

%% file: contents/abstract.tex
Minimizing response times is crucial for emergency medical services to reduce patients' waiting times and to increase their survival rates. 
Many models exist to optimize operational tasks such as ambulance allocation and dispatching. Including accurate demand forecasts in such models can improve operational decision-making. Against this background, we present a novel convolutional neural network (CNN) architecture that transforms time series data into heatmaps to predict ambulance demand. Applying such predictions requires incorporating external features that influence ambulance demands. We contribute to the existing literature by providing a flexible, generic CNN architecture, allowing for the inclusion of external features with varying dimensions. Additionally, we provide a feature selection and hyperparameter optimization framework utilizing Bayesian optimization. We integrate historical ambulance demand and external information such as weather, events, holidays, and time. To show the superiority of the developed CNN architecture over existing approaches, we conduct a case study for Seattle’s 911 call data and include external information. We show that the developed CNN architecture outperforms existing state-of-the-art methods and industry practice by more than 9\%.

%% file: contents/introduction.tex
\section{Introduction}
Reducing response times is paramount for emergency medical services (EMS) to provide first aid in a timely manner. However, tight budgets pressure EMS systems to minimize operational expenses, resulting in limited availability of resources. Demographic changes such as an aging population can further increase ambulance demands, intensifying the need for an efficient use of resources. Climate change causing more frequent extreme weather conditions, such as heat spells or heavy rainfalls, can additionally challenge EMS systems in the future. To tackle these challenges and to reduce response times, several models exist to optimize operational tasks such as ambulance allocation and dispatching, see, e.g., \cite{brotcorne2003ambulance}, \cite{aboueljinane2013review}, \cite{belanger2019recent}, and \cite{farahani2019or} for literature overviews. 
Embedding accurate demand forecasts in such models can significantly improve operational decision-making. 

In literature, various machine learning approaches such as artificial neural networks (ANNs), tree-based models, or support vector regressions (SVRs) exist to predict ambulance demand. For other prediction tasks, such as traffic data and mobility demand predictions, convolutional neural networks (CNNs) yield highly accurate predictions, see, e.g., \cite{wang2018deepstcl} and \cite{guo2019deep}. Although CNNs can outperform state-of-the-art approaches for such prediction tasks, CNNs have not yet been applied for ambulance demand prediction. Against this background, we present a novel CNN architecture that transforms time series information into heatmaps to forecast ambulance demand.

 Our contribution to the existing literature is three-fold: First, we present a flexible, generic CNN architecture, allowing for the inclusion of external features with varying dimensions. This concept differs from most state-of-the-art approaches that divide the given region into subregions and derive local forecasts for each of them separately, neglecting possible spatial correlations. Contrarily, we enable the detection of correlations in space and time by including three-dimensional convolutional layers in a CNN architecture and obtain the predictions for all subregions simultaneously. Most CNN architectures neglect external information as they have been mainly applied to image classification and recognition tasks. Nonetheless, effectively predicting ambulance demand with CNN architectures requires the incorporation of external features. Thus, we integrate historical ambulance demand and external information such as weather, events, holidays, time, weekdays, and months.
Second, we show how to jointly perform feature selection and hyperparameter tuning by applying Bayesian optimization (BO). We treat the decision of whether to include a feature or not as an additional hyperparameter of the prediction model and add these parameters to the hyperparameter tuning process. 
To tackle the high-dimensional search space, we introduce a novel hierarchical BO approach and apply BO with dimension dropout. 
Third, we analyze the features' importance by calculating the SHapley Additive exPlanation (SHAP) values representing the contribution of each feature to the prediction. Additionally, we explore the CNN's performance for different forecasting horizons by conducting a sensitivity analysis. 

We base our results on a case study of Seattle’s 911 call data and include external information. Results show that the developed CNN outperforms state-of-the art methods and industry practice by more than 9\%. Incorporating feature selection in the BO reduces the number of parameters by 40.4\% while the performance decreases by less than 0.05\%. The SHAP values show that the CNN is more robust to changes in the historical ambulance demand, compared to its benchmarks. This enables a better handling of demand outliers or data inconsistencies. We further show that the BO with dimension dropout and the hierarchical BO outperform basic BO approaches and random search.

The remainder of this paper is structured as follows. We first present related literature in Section \ref{sec:literature}. Section 3 describes our problem setting. Section \ref{sec:methodology} introduces the CNN architecture and our algorithmic framework to conduct feature selection and hyperparameter tuning. In Section \ref{sec:castestudy}, we introduce the numerical case study for Seattle’s 911 call data and present our results in Section \ref{sec:experimentalresults}. Section \ref{sec:conclusion} concludes this work.

%% file: contents/relatedwork.tex
\section{Related work}\label{sec:literature}

Many early but also recent studies apply regression models to predict ambulance demand, investigating the influence of socio-economic variables, e.g., population, age, gender, race, land use, living conditions, education, employment, income, or marital status, see, e.g., \cite{aldrich1971analysis}, \cite{siler1975predicting},
\cite{lowthian2011challenges}, and \cite{steins2019forecasting}.
\cite{wong2020effects} apply multivariate forward regression and show that EMS demand for specific patient groups, such as elderly or critical patients, is sensitive to weather conditions and changes throughout the day. 
Another stream of literature focuses on time series models to predict ambulance demand. 
\cite{baker1986determination} apply Winter's exponential smoothing \citep{winters1960forecasting} and present a goal programming approach to optimize the weightings given to different forecasting statistics. 
\cite{tandberg1998time} implement different time series models to predict hourly emergency incidents in Albuquerque and show that simple time series models can outperform more expensive, complex models. 
\cite{channouf2007application} include different day-of-week, month-of-year, and special-day effects in their time series models. They show that considering historic call volumes of previous hours can improve the hourly forecast accuracy. 
\cite{matteson2011forecasting} make predictions on an hourly level and introduce a factor model combined with time series models. They include time information such as the weekday and week. 
\cite{vile2012predicting} present a singular spectrum analysis to forecast emergency calls in Wales and show that their approach outperforms Holt-Winters and autoregressive integrated moving-average (ARIMA) models for long-term forecasts. 
\cite{wong2014weather} integrate weather data in an ARIMA model and show that the prediction error of the 7-day forecast can be reduced by 10\% by incorporating weather data in the model. \cite{gijo2016sarima} introduce a Seasonal ARIMA model to predict the emergency call volume for a state in India.
\cite{nicoletta2016bayesian} introduce a Bayesian approach, modeling ambulance demand as a generalized linear mixed model. Results show that the time-of-day, holidays, each zone's population, area, and type heavily influence the prediction's performance. 
\cite{zhou2015spatio} present a Gaussian mixture model to predict ambulance demand in Toronto, Canada, tackling data sparsity caused by fine temporal resolutions. \cite{zhou2016predicting} apply kernel density estimation and introduce a kernel warping approach, a form of Laplacian eigenmaps belonging to the field of manifold learning. 
\cite{setzler2009ems} present a multilayer perceptron (MLP) including four categorical variables: time-of-day, day-of-week, month, and season. 
\cite{chen2015demand} apply an ANN, moving average, sinu-soidal regression, and support vector regression model for ambulance demand prediction. For each subregion, they select the best-performing model.
\cite{wang2021forecasting} include a heterogeneous multi-graph convolutional layer in a neural network to predict ambulance demand. To form the graph, they make use of dispatching areas. 
\cite{jin2021predicting} introduce a bipartite graph convolutional network to predict ambulance demand for region-hospital pairs. They include regional features and hospital features, e.g., capacity information.

In summary, many forecasting approaches exist to predict ambulance demand. These approaches include regression analyses, time series models, mixture models, ANNs, graph-based models, SVRs, and tree-based models. We provide a summary including the features considered in each study in Table \ref{table:literatureAnalysis} and refer to \cite{reuter2017logistics} for an overview. In the field of mobility demand forecasting, few approaches exist that base forecasting on CNNs, which have so far mostly been used for image classification, to implicitly capture spatio-temporal correlations \citep{wang2018deepstcl,guo2019deep}. Although CNNs yield highly accurate predictions for such applications, they have not yet been applied to predict ambulance demand. Moreover, a framework that allows generic feature integration, efficient feature selection, and hyperparameter tuning for such architectures has neither been developed for general mobility demand nor for ambulance demand forecasting. To close this research gap, we introduce a novel CNN architecture for predicting ambulance demand that is able to generically incorporate external features and embed it in an efficient hyperparameter tuning framework.

\input{Tables/LiteratureAnalysis}

%% file: Tables/LiteratureAnalysis.tex
\begin{table}[h!]

\Huge
\centering
\begin{spacing}{1.4}
\begin{adjustbox}{width=.89\textwidth}
\begin{tabular}{l l c c c c c c c c c c c c c c c c c c c c c c c c c c c c c c c c c c c c}

&&\rotatebox{90}{Age}&\rotatebox{90}{Area Information (e.g. Size)}&\rotatebox{90}{Baromatic Pressure}&\rotatebox{90}{Clouds}&\rotatebox{90}{Crime Rate}&\rotatebox{90}{Day of the Month}&\rotatebox{90}{Day of the Week / Weekend}&\rotatebox{90}{Disaster Risk Index}&\rotatebox{90}{Education}&\rotatebox{90}{Employment}&\rotatebox{90}{EMS Characteristics}&\rotatebox{90}{Events}&\rotatebox{90}{Gender}&\rotatebox{90}{Health Care Infrastructure}&\rotatebox{90}{Historic Data}&\rotatebox{90}{Holidays}&\rotatebox{90}{Hour / Intraday Pattern}&\rotatebox{90}{Humidity}&\rotatebox{90}{Income / Wealth}&\rotatebox{90}{Land Use \& Housing}&\rotatebox{90}{Living Conditions}&\rotatebox{90}{Marital Status}&\rotatebox{90}{Mobility}&\rotatebox{90}{Week / Month / Season of Year}&\rotatebox{90}{Non-Resident Population}&\rotatebox{90}{Population}&\rotatebox{90}{Racial Traits / Birthplace}&\rotatebox{90}{Rain / Precipitation}&\rotatebox{90}{Mobility Infrastructure}&\rotatebox{90}{Temperature}&\rotatebox{90}{Traffic }&\rotatebox{90}{Visibility}&\rotatebox{90}{Wind speed}&\rotatebox{90}{Year}\\\hline

\multirow{13}{*}{\rotatebox{0}{\shortstack[l]{Regr.\\Models}}}
&[1]&\checkmark&\checkmark&-&-&-&-&-&-&-&\checkmark&\checkmark&-&\checkmark&-&-&-&-&-&\checkmark&\checkmark&\checkmark&\checkmark&-&-&-&\checkmark&\checkmark&-&\checkmark&-&-&-&-&-\\
&[2]&\checkmark&-&-&-&-&-&-&-&\checkmark&-&-&-&-&-&-&-&-&-&\checkmark&-&\checkmark&-&-&-&-&\checkmark&\checkmark&-&-&-&-&-&-&-\\
&[3]&-&-&-&-&-&-&-&-&-&\checkmark&-&-&\checkmark&-&-&-&-&-&-&\checkmark&-&\checkmark&-&-&-&-&\checkmark&-&-&-&-&-&-&-\\
&[4]&\checkmark&\checkmark&-&-&-&-&-&-&-&\checkmark&-&-&\checkmark&-&-&-&-&-&\checkmark&\checkmark&-&-&\checkmark&-&\checkmark&-&\checkmark&-&\checkmark&-&-&-&-&-\\
&[5]&\checkmark&-&-&-&-&-&-&-&\checkmark&\checkmark&\checkmark&-&\checkmark&-&-&-&-&-&\checkmark&\checkmark&\checkmark&\checkmark&\checkmark&-&-&\checkmark&\checkmark&-&-&-&-&-&-&-\\
&[6]&-&-&-&-&-&-&\checkmark&-&-&-&-&-&-&-&\checkmark&-&-&-&-&-&-&-&-&-&-&-&-&-&-&-&-&-&-&-\\
&[7]&\checkmark&-&-&-&-&-&-&-&-&-&-&-&-&-&-&-&-&-&\checkmark&-&-&-&-&-&\checkmark&\checkmark&-&-&\checkmark&-&-&-&-&-\\
&[8]&\checkmark&-&-&-&-&-&-&-&\checkmark&-&-&-&\checkmark&\checkmark&-&-&-&-&\checkmark&\checkmark&\checkmark&-&-&-&-&-&-&-&-&-&-&-&-&-\\
&[9]&\checkmark&-&-&-&-&-&-&-&\checkmark&\checkmark&-&-&-&-&-&-&-&-&-&\checkmark&-&-&-&-&-&-&-&-&-&-&-&-&-&-\\
&[10]&\checkmark&-&-&-&-&-&-&-&-&-&-&-&\checkmark&-&-&-&-&-&-&-&-&-&-&-&-&-&-&-&-&-&-&-&-&-\\
&[11]&-&-&-&-&-&\checkmark&\checkmark&-&-&-&-&-&-&-&\checkmark&-&\checkmark&-&-&-&-&-&-&\checkmark&-&-&-&\checkmark&-&-&\checkmark&-&-&\checkmark\\
&[12]&\checkmark&-&-&-&-&-&-&-&-&\checkmark&-&-&-&-&-&-&-&-&\checkmark&-&\checkmark&-&-&-&-&\checkmark&\checkmark&-&\checkmark&-&-&-&-&-\\
&[13]&-&-&\checkmark&\checkmark&-&-&\checkmark&-&-&-&-&-&-&-&-&\checkmark&-&\checkmark&-&-&-&-&-&-&-&-&-&\checkmark&-&\checkmark&-&\checkmark&\checkmark&-\\\hline

\multirow{8}{*}{\rotatebox{0}{\shortstack[l]{Time\\Series\\Models}}}
&[6]&-&-&-&-&-&-&-&-&-&-&-&-&-&-&\checkmark&-&-&-&-&-&-&-&-&-&-&-&-&-&-&-&-&-&-&-\\
&[14]&-&-&-&-&-&-&-&-&-&-&-&-&-&-&\checkmark&-&-&-&-&-&-&-&-&-&-&-&-&-&-&-&-&-&-&-\\
&[15]&-&-&-&-&-&-&\checkmark&-&-&-&-&\checkmark&-&-&\checkmark&-&-&-&-&-&-&-&-&\checkmark&-&-&-&-&-&-&-&-&-&-\\
&[16]&-&-&-&-&-&-&\checkmark&-&-&-&-&-&-&-&\checkmark&-&\checkmark&-&-&-&-&-&-&\checkmark&-&-&-&-&-&-&-&-&-&-\\
&[17]&-&-&-&-&-&-&-&-&-&-&-&-&-&-&\checkmark&-&-&-&-&-&-&-&-&-&-&-&-&-&-&-&-&-&-&-\\
&[18]&-&-&-&-&-&-&-&-&-&-&-&-&-&-&\checkmark&-&-&\checkmark&-&-&-&-&-&-&-&-&-&-&-&\checkmark&-&-&-&-\\
&[19]&-&-&-&-&-&-&-&-&-&-&-&-&-&-&\checkmark&-&-&-&-&-&-&-&-&-&-&-&-&-&-&-&-&-&-&-\\
&[11]&-&-&-&-&-&\checkmark&\checkmark&-&-&-&-&-&-&-&\checkmark&-&\checkmark&-&-&-&-&-&-&\checkmark&-&-&-&\checkmark&-&-&\checkmark&-&-&\checkmark\\\hline

\multirow{3}{*}{\rotatebox{0}{\shortstack[l]{Mixt.\\Models}}}
&[20]&-&-&-&-&-&-&\checkmark&-&-&-&-&-&-&-&\checkmark&-&\checkmark&-&-&-&-&-&-&-&-&-&-&\checkmark&-&\checkmark&-&-&-&-\\
&[21]&-&\checkmark&-&-&-&-&-&-&-&-&-&-&-&-&-&\checkmark&\checkmark&-&-&\checkmark&-&-&-&-&-&\checkmark&-&-&-&-&-&-&-&-\\\hline

\rotatebox{0}{KDEs}
&[22]&-&-&-&-&-&-&-&-&-&-&-&-&-&-&\checkmark&-&-&-&-&-&-&-&-&-&-&-&-&-&-&-&-&-&-&-\\\hline

\multirow{2}{*}{\rotatebox{0}{ANNs}}
&[23]&-&-&-&-&-&-&\checkmark&-&-&-&-&-&-&-&-&-&\checkmark&-&-&-&-&-&-&\checkmark&-&-&-&-&-&-&-&-&-&-\\
&[11]&-&-&-&-&-&\checkmark&\checkmark&-&-&-&-&-&-&-&\checkmark&-&\checkmark&-&-&-&-&-&-&\checkmark&-&-&-&\checkmark&-&-&\checkmark&-&-&\checkmark\\\hline

\multirow{2}{*}{\rotatebox{0}{\shortstack[l]{Graph\\-based\\Models}}}
&[24]&-&-&-&-&\checkmark&-&-&-&-&-&-&-&-&\checkmark&\checkmark&-&-&-&-&\checkmark&-&-&-&-&\checkmark&\checkmark&-&-&-&-&-&-&-&-\\
&[25]&-&\checkmark&-&-&-&-&-&\checkmark&-&-&-&-&-&-&\checkmark&-&-&-&-&-&-&-&\checkmark&-&\checkmark&\checkmark&-&-&\checkmark&-&-&-&-&-\\\hline

\rotatebox{0}{SVR}
&[11]&-&-&-&-&-&\checkmark&\checkmark&-&-&-&-&-&-&-&\checkmark&-&\checkmark&-&-&-&-&-&-&\checkmark&-&-&-&\checkmark&-&-&\checkmark&-&-&\checkmark\\\hline

\rotatebox{0}{Trees}
&[11]&-&-&-&-&-&\checkmark&\checkmark&-&-&-&-&-&-&-&\checkmark&-&\checkmark&-&-&-&-&-&-&\checkmark&-&-&-&\checkmark&-&-&\checkmark&-&-&\checkmark\\\hline

\end{tabular}
\end{adjustbox}
\begin{spacing}{0.8}
\caption{\centering{Literature Analysis
\textmd{(
[1] \cite{aldrich1971analysis},
[2] \cite{gibson1971analysis},
[3] \cite{siler1975predicting},
[4] \cite{kvaalseth1979statistical},
[5] \cite{kamenetzky1982estimating},
[6] \cite{baker1986determination},
[7] \cite{cadigan1989predicting},
[8] \cite{svenson2000patterns},
[9] \cite{cramer2012predicting},
[10] \cite{lowthian2011challenges},
[11] \cite{chen2015demand},
[12] \cite{steins2019forecasting},
[13] \cite{wong2020effects},
[6] \cite{baker1986determination},
[14] \cite{tandberg1998time},
[15] \cite{channouf2007application},
[16] \cite{matteson2011forecasting},
[17] \cite{vile2012predicting},
[18] \cite{wong2014weather},
[19] \cite{gijo2016sarima},
[11] \cite{chen2015demand},
[20] \cite{zhou2015spatio},
[21] \cite{nicoletta2016bayesian},
[22] \cite{zhou2016predicting},
[23] \cite{setzler2009ems},
[11] \cite{chen2015demand},
[24] \cite{jin2021predicting},
[25] \cite{wang2021forecasting},
[11] \cite{chen2015demand},
[11] \cite{chen2015demand})
}
}}
\label{table:literatureAnalysis}
\end{spacing}
\end{spacing}

\end{table}

%% file: contents/problemsetting.tex
\section{Problem Setting}\label{sec:problem_description}

We divide the region for which we are making the forecast 
into $q\times p$ subregions and aim at predicting the ambulance demand for all subregions for a time period $t$. For each subregion, the ambulance demand can be represented as discrete time series $z \in \mathbb{R}^{T}$.

{\definition[Time series]{Let $z \in \mathbb{R}^{T}$ be a discrete time series containing a sequence of $T$ observations sampled from a random variable $\mathcal{Z}$ at equidistant timesteps. 
}}\vspace*{0.5\bigskipamount}

Based on each time series, conventional algorithms, e.g., exponential smoothing \citep{baker1986determination} or ARIMA models \citep{tandberg1998time}, can predict the future ambulance demand for every subregion in an iterative fashion. However, such iterative approaches neglect spatial correlations between the time series. For this reason, we aim at generating an integrated forecast for all subregions simultaneously. To enable an integrated forecast, we represent the ambulance demand as a spatio-temporal time series assuming a spatial correlation between the subregions’ ambulance demands.

{\definition[Spatio-temporal time series]{Assuming a spatial correlation between the subregions' time series, we can represent the ambulance demand as a $q \times p$ matrix of spatially correlated time series, each of length T, which we denote by $Z \in \mathbb{R}^{q \times p \times T}$.}}\vspace*{0.5\bigskipamount}

In addition, effectively predicting ambulance demand requires the incorporation of external features. In the existing literature, various external features have been considered such as socio-economic or weather data. These external features strongly differ in their properties. For example, socio-economic data mostly differs across city districts but only changes gradually over time. In contrast, weather data can change quickly but does not strongly differ across city districts. To generically include external features, we distinguish four types: 

\begin{description}
\item[Three-dimensional input data.] Let $X^{\text{3D}}=\{{X_{i}^{\text{3D}}}\}_{i=1}^{n}$ be the three-dimensional features of our data set, where each data instance ${X^{\text{3D}}_i} \in \mathbb{R}^{q \times p \times L}$ is a spatio-temporal time series representing the historic ambulance demand of $L$ periods for all $q \times p$ subregions. 

\item[Two-dimensional input data.] Let $X^{\text{2D}}=\{{X_{i}^{\text{2D}}}\}_{i=1}^{n}$ be the two-dimensional features of our data set, where each data instance ${X_{i}^{\text{2D}}} = \{{x_{i_{j}}^{\text{2D}}}\}_{j=1}^{m}$ is a set of $m$ two-dimensional vectors, i.e., ${x_{{i}_j}^{\text{2D}}} \in \mathbb{R}^{q \times p}$. We include two-dimensional data in the case that only the spatial dimension is considered, e.g., for representing events taking place in the predicted period.

\item[One-dimensional input data.] 

Let $X^{\text{1D}}=\{{X_{i}^{\text{1D}}}\}_{i=1}^{n}$ be the one-dimensional features of our data set, where each data instance ${X_{i}^{\text{1D}}} = \{{X_{i_{j}}^{\text{1D}}}\}_{j=1}^{r}$ is a set of $r$ one-dimensional vectors, i.e., ${x_{{i}_j}^{\text{1D}}} \in \mathbb{R}^{l_j}$, where $l_j$ is the length of vector $j$. One-dimensional features are either time series such that $l_j=L$, where $L$ is the number of past periods to be considered, or one-hot-encoded information, e.g., for including time information such as the month, weekday, or hour.

\item[Scalar input data.] Let $X^{\text{S}}=\{{X_{i}^{\text{S}}}\}_{i=1}^{n}$ be the scalar features of our data set, where each data instance ${X_{i}^{\text{S}}} = \{{x_{i_{j}}^{S}}\}_{j=1}^{s}$ is a set of $s$ scalar features, i.e., ${x_{{i}_j}^{\text{S}}} \in \mathbb{R}$. We include scalar inputs for data without spatial or temporal distribution which is not one-hot-encoded, e.g., the temperature prediction for period $t$ for which we neglect the spatial distribution.
\end{description}

Let $(\textbf{X}, \textbf{y}) = \{\textbf{x}_i, y_i\}_{i=1}^{n}$ be our dataset. $\textbf{x}_i = \{ X^{\text{1D}}_i, X^{\text{2D}}_i, X^{\text{3D}}_i, X^{\text{S}}_i \} \in \mathbb{R}^{v}$ represents a $v$-dimensional feature vector and $y_i \in \mathbb{R}$ its respective dependent variable. Each data instance is independently sampled from an unknown distribution $(\mathcal{X}, \mathcal{Y})$. Our model $\mathcal{M}$ aims at learning a function $F_{\mathcal{M}}:\mathcal{X}\rightarrow\mathcal{Y}$ that maps each input vector $x_i$ to its associated variable $y_i$.

%% file: contents/methodology.tex
\section{Methodology}\label{sec:methodology}

We first introduce our CNN architecture in Section \ref{sec:methodology_cnn}. Then, we describe our BO approaches to perform hyperparameter optimization and feature selection in Section \ref{sec:hyperparamteropt_featuresel}. 

\subsection{Convolutional Neural Network Architecture}\label{sec:methodology_cnn}

To predict ambulance demand, we apply a novel CNN architecture. Leveraging our neural network with convolutional layers enables the generation of feature maps, i.e., transformed images, highlighting patterns identified in the input image. To form these feature maps, filters, i.e., weight matrices, traverse the input image with a defined stride, i.e., step size, and conduct convolutions to detect patterns that the filters have learned to identify. So far, CNNs have mainly been used to solve image classification tasks, in which convolutional layers detect spatial patterns in the images. We apply this concept to reveal spatio-temporal patterns within our historic ambulance demand. We further integrate external features through concatenation at different stages of the CNN. We visualize our CNN architecture in Figure \ref{fig:cnn_architecture} and provide a more detailed overview in Appendix \ref{sec:cnn_architecture}.

\begin{figure}
  \includegraphics[width=\linewidth]{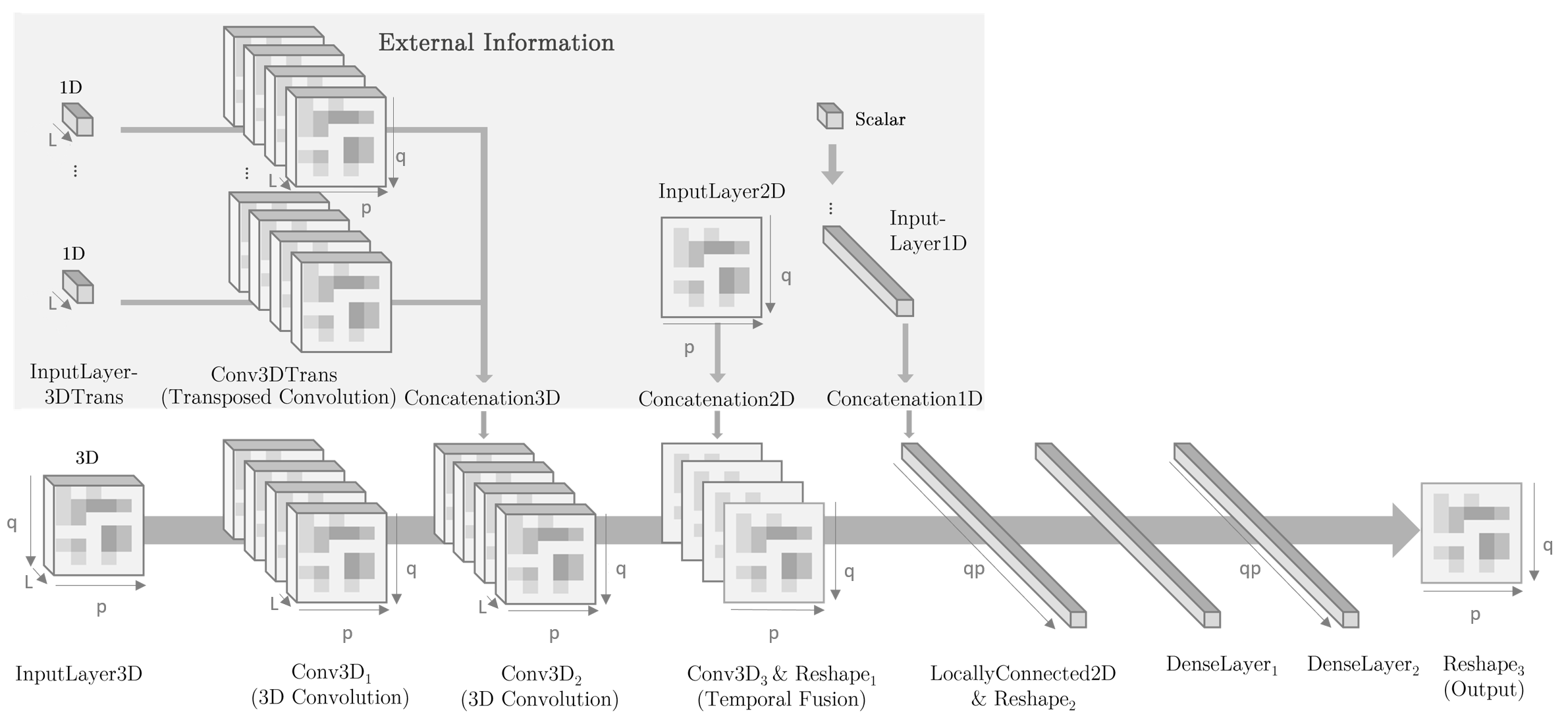}
  \caption{Visualization: CNN Architecture}
  \label{fig:cnn_architecture}
\end{figure}

We apply three-dimensional convolutional layers to detect spatio-temporal patterns in the historic ambulance demand, represented as spatio-temporal time series $X^{3D}\in\mathbb{R}^{q \times p \times L}$, where $L$ is the number of past periods that we take into account. 
Each convolutional layer $l$ contains $M_l$ filters. The weight of filter $h$ in layer $l$ at position $(w_1,w_2,w_3)$ is $w_{h}^l(w_1,w_2,w_3)$. We define the filter size by $k^l=(k_{1}^l,k_{2}^l,k_{3}^l)$ and apply a stride of $s^l=(s_{1}^l,s_{2}^l,s_{3}^l)$. We denote the activation function in layer $l$ by $\vartheta^l(\cdot)$ and the bias by $b_{h}^{l}$. By applying zero padding, i.e., surrounding the input image with additional zero-valued pixels, we maintain the output's size throughout consecutive layers. The coordinates $(x_1,x_2,x_3)$ uniquely define the values within each feature map, assuming zero-based indexing. We calculate the number of parameters $\phi_l$ learned for layer $l$ by

\begin{equation}
\phi_l = (k_{1}^{l}k_{2}^{l}k_{3}^{l}M_{l-1}+1)M_l,\label{eq:nrparam3dconv}
\end{equation}

and derive the output of filter $h$ in layer $l$ by
 
\begin{equation}
\begin{split}
o_{h}^{l}\left(x_1,x_2,x_3\right)=\vartheta^l\Bigg(\sum_{m=0}^{M_{l-1}-1}\sum_{w_1=0}^{k_{1}-1}\sum_{w_2=0}^{k_{2}-1}\sum_{w_3=0}^{k_{3}-1}
&o_{m}^{l-1}\left(s_{1}^{l}x_1+w_1, s_{2}^{l}x_2+w_2, s_{3}^{l}x_3+w_3\right)\\&*w_{h}^{l}\left(w_1,w_2,w_3\right)+b_{h}^{l}\Bigg).\label{eq:3dconv}
\end{split}
\end{equation}
 
We further include transposed convolutions in the model in order to upsample one-dimensional input data to three dimensions. This enables us to weigh the impact of one-dimensional data spatially and, simultaneously, take temporal changes into account. 
For example, we assume that the area's size allows the negligence of spatial temperature differences. However, \cite{wong2020effects} show that temporal temperature changes influence peoples' health conditions. Furthermore, extreme weather can influence specific areas within the depicted region more severely, such as districts with an elderly population during hot spells or urban watersides in the case of heavy rainfalls. 
We calculate the output of the transposed convolution of filter $h$ in layer $l$ as follows

\begin{equation}
\begin{split}
\label{eq:transposedconv}
o_{h}^{l}\left(x_1, x_2, x_3\right)
=\vartheta^l
\Bigg(
\sum_{m=0}^{M_{l-1}-1}
&\sum_{\substack{i_1=0,\\\iota(x_1,i_1,s_{1}^{l})\\<k_1}}^{\left\lfloor{\frac{x_1}{s_{1}^{l}}}\right\rfloor}
\sum_{\substack{i_2=0,\\\iota(x_2,i_2,s_{2}^{l})\\<k_2}}^{\left\lfloor{\frac{x_2}{s_{2}^{l}}}\right\rfloor}\sum_{\substack{i_3=0,\\\iota(x_3,i_3,s_{3}^{l})\\<k_3}}^{\left\lfloor{\frac{x_3}{s_{3}^{l}}}\right\rfloor}
o_{m}^{l-1}
\left(
{\left\lfloor{\frac{x_1}{s_{1}^{l}}}\right\rfloor-i_1},
{\left\lfloor{\frac{x_2}{s_{2}^{l}}}\right\rfloor-i_2},
{\left\lfloor{\frac{x_3}{s_{3}^{l}}}\right\rfloor-i_3}
\right)\\
&*w_{h}^{l}\left(
{\iota\left(x_1,i_1,s_1\right)},
{\iota\left(x_2,i_2,s_2\right)},
{\iota\left(x_3,i_3,s_3\right)}
\right)
+b_{h}^{l}
\Bigg),
\end{split}
\end{equation}
where
\begin{align}
\iota\left(x,i,s\right)=x-s\left(\left\lfloor{\frac{x}{s}}\right\rfloor-i\right).
\end{align}

After conducting the convolutions (\ref{eq:3dconv}) and (\ref{eq:transposedconv}), we concatenate the generated feature maps. To enable this concatenation, we set the stride, padding, and filter sizes such that all feature maps are of shape $q\times p \times L$. We then fuse the concatenated feature maps along the temporal dimension. We concatenate the temporal fusion's outputs, i.e., two-dimensional feature maps of shape $q\times p$, with the two-dimensional input data. In the following, we add a locally connected layer to combine the identified patterns with the included inputs. Here, we substitute $w^{l}_{h}$ by $w^{l}_{hx_1x_2}$ and $b^{l}_{h}$ by $b^{l}_{hx_1x_2}$ to represent the weights and bias in filter $h$ at position $(x_1,x_2)$. 
We compute the outputs as follows

\begin{equation}
\begin{split}
o_{h}^{l}\left(x_1,x_2\right)=\vartheta^l\Bigg(\sum_{m=0}^{M_{l-1}-1}&\sum_{w_1=0}^{k_{1}-1}\sum_{w_2=0}^{k_{2}-1}
o_{m}^{l-1}\left(x_{1}s_{1}^{l}+w_1, x_{2}s_{2}^{l}+w_2\right)\\&*w_{hx_1x_2}^{l}\left(w_1,w_2\right)+b_{hx_1x_2}^{l}\Bigg). \label{eq:fullyconnected}
\end{split}
\end{equation}

For the locally connected layer $l$, we train $\phi_l$ parameters calculated by
\begin{equation}
\phi_l = (M_{l-1}+1)qp,\label{eq:nrparamlocconn}
\end{equation}
where $qp$ is the number of output neurons.

We then embed one-dimensional data and scalar inputs through concatenation. Finally, we add two dense layers to enable the network to learn from these inputs. Each dense layer $l$ consists of $N^{l}$ neurons, each denoted by $v^{l}_{i}$ where $i=1,...,N^{l}$. The activation function and bias for layer $l$ are $\vartheta^{l}\left(\cdot\right)$ and $b^{l}$. We denote the weights between neurons $v^{l}_{i}$ and $v^{l-1}_{j}$ by $w^{l}_{ij}$. Thus, we calculate the output $a^{l}_{i} 
$ of neuron $v^{l}_{i} 
$ as follows
\begin{align}
    a^{l}_{i}=\vartheta^{l}\Big(\sum_{j=1}^{N^{l-1}}w^{l}_{ij}a^{l-1}_{j}+b^{l}\Big).\label{eq:denselayer}
\end{align}

We calculate the number of learned parameters for dense layer $l$ by
\begin{align}
    \phi_l=(N^{l-1}+1)N^{l}.\label{eq:nrparamdense}
\end{align}

In the last layer, we set the number of neurons to $qp$ such that we can reshape its outputs into a $q \times p$ heatmap, constituting the model's prediction. 

\subsection{Hyperparameter Optimization and Feature Selection}\label{sec:hyperparamteropt_featuresel}

To provide a lean model and to improve the model's performance, we conduct feature selection and tune the model's hyperparameters. Adding or eliminating features may change the network architecture, as some input layers are only added when including certain external information. 
Thus, optimizing hyperparameters before conducting feature selection may lead to a suboptimal choice of parameters, e.g., the number of filters and their sizes may depend on the features selected. Similarly, conducting feature selection before hyperparameter tuning may lead to suboptimal results, since the selection process may be based on a network with insufficiently tuned hyperparameters. For this reason, we aim at utilizing an intrinsic feature selection method tuning the CNN's hyperparameters and conducting feature selection simultaneously.
 
 Therefore, we treat the decision of whether to include a feature or not as an additional hyperparameter of the prediction model and include these parameters in the hyperparameter tuning process. Pursuing such an approach entails the following challenges: First, the model has numerical and categorical hyperparameters that may depend on each other, e.g., adding layers requires their hyperparameters to be tuned. Vice versa, when removing layers, we can neglect their hyperparameters in the tuning process. Second, conducting feature selection for $N$ features increases the search space by $2^N$, resulting in a high-dimensional search space. Third, evaluating the predictive model is computationally expensive. Accordingly, common hyperparameter optimization strategies, such as manual or grid search, are unsuitable for adequately covering the search space in an acceptable amount of computational time. Fourth, the stochastic nature of the training process entails noise, as similar hyperparameter values can result in different function values. Finally, derivatives of the underlying target function are not given, and multiple local optima may exist due to the target function's non-convexity. To tackle these challenges and to learn from information gained by previously performed parameter combinations, we apply BO. 

\paragraph{Bayesian Optimization} BO is a sequential model-based optimization (SMBO) approach that is used to optimize the hyperparameters of a black-box function $f\left(\cdot\right)$, which is expensive to evaluate. We leverage a surrogate model to approximate the target function to bypass its expensive evaluation. In each iteration of the BO, we update the surrogate model based on our previous observations. Then, we optimize an acquisition function utilizing the surrogate model, which derives the next promising hyperparameter settings to be evaluated. 

We denote $J$ hyperparameters by $\theta=\lambda_1, \lambda_2, \lambda_3, ... ,\lambda_J$ and the search space by $\Theta=\Lambda_1 \times \Lambda_2 \times \Lambda_3 \times ... \times \Lambda_J$. Then, we aim to solve the following optimization problem:
\begin{align}
\theta^{*}&= \arg\min_{\theta\in\Theta} f\left(\theta\right).
\end{align}

As the target function $f\left(\theta\right)$ is computationally expensive to evaluate, we model this function via the surrogate model $\mathcal{M}$. As surrogates, we apply Gaussian processes \citep{snoek2012practical}, random forests \citep{hutter2011sequential}, and extremely randomized trees \citep{geurts2006extremely}. In each iteration $\iota$ of the BO, we optimize the acquisition function $\mathcal{A}$ to derive a new promising parameter setting as follows
\begin{align}
    \theta_{\iota} \gets \arg\max_{\theta\in\Theta} \mathcal{A}(\theta).
\end{align}

Here, we apply an \textit{Expected Improvement} strategy enabling a trade-off between exploration and exploitation: We define the best (minimal) observed function value of $n$ iterations by $f^{*}=\text{min}(f(\theta_{1}),...,f(\theta_{n}))$. When sampling from $f$ at a new, unknown point, we treat its realization $Y$ as a normally distributed variable $Y\sim\mathcal{N}(\mu,\sigma^2)$. Thus, the expected improvement at point $\theta$ is
\begin{align}
\mathbb{E}[\mathbf{I}(\theta)]=\mathbb{E}[\text{max}(f^*-Y,0)]=(f^*-\mu)\Phi\left(\frac{f^*-\mu}{\sigma}\right)+\sigma\phi\left(\frac{f^*-\mu}{\sigma}\right),
\end{align}

where $\phi(\cdot)$ and $\Phi(\cdot)$ are the standard normal density and distribution functions, correspondingly \citep{jones1998efficient}. 

We present a basic pseudo-code for BO in Algorithm \ref{alg:BO}. After initializing the set of observations $R=\{(\theta_{1},f(\theta_{1})),(\theta_{2},f(\theta_{2})),...,(\theta_{m},f(\theta_{m}))\}$ via $m$ iterations of random search (l. 3), we conduct $n$ iterations of BO. In each iteration $\iota$, we update the surrogate model and optimize the acquisition function $\mathcal{A}$ (l. 6). We then evaluate the target function $f$ at the next promising point $\theta_{\iota}$ and update the set of results $R$ with our new observation (l. 7). Finally, we return the incumbent parameter settings $\theta^{*}$ (l. 9).
\input{Algorithms/alg1}

\paragraph{High-Dimensional Bayesian Optimization} High-dimensional search spaces challenge existing BO methods. Although classic approaches are still applied, e.g., random search \citep{bergstra2012random} or BO with random forests \citep{hutter2011sequential}, more advanced methods have been developed to tackle high-dimensional search spaces. For example, \cite{wang2013bayesian} make use of low effective dimensions, i.e., dimensions which do not significantly change the objective function, to reduce the search space. However, assuming that only a subset of dimensions is effective may not be possible, e.g., if all dimensions highly influence the objective. 
To handle problems with high effective dimensionality, \cite{li2018high} introduce the concept of random dimension dropout, which optimizes the acquisition function only over a subset of dimensions. In this study, we implement two different approaches. First, we build upon \cite{li2018high} and apply BO with dimension dropout. Second, we present a novel hierarchical BO to reduce the search space and optimize hyperparameters sequentially. 
For comparison, we refer to the previously introduced BO approaches without dimension dropout as \textit{basic} BO approaches.

\paragraph{A. Bayesian Optimization with Dimension Dropout}
Based on \cite{li2018high}, we implement BO with dimension dropout as detailed in Algorithm \ref{alg:BO_dropout}.
\input{Algorithms/alg2.tex}
 First, we derive $m$ initial points via random search and add them to the set of observations $R=\{(\theta_{1},f(\theta_{1})),(\theta_{2},f(\theta_{2})),...,(\theta_{m},f(\theta_{m}))\}$ (l. 3). Then, we conduct $n$ iterations of BO with dimension dropout. In each iteration $\iota$, we randomly draw $d$ dimensions out of the problem's search space dimensions $\mathcal{D}$ and refer to the subset of drawn dimensions as $D'$. We ensure that $d=|D'|<|\mathcal{D}|$. To dynamically adapt $d$ in accordance with the total number of dimensions, we calculate $d$ by $d=\left\lfloor\Tilde{d}*|\mathcal{D}|\right\rfloor$, where $\Tilde{d}\in[0,1]$ (l. 6). 
We denote the parameters of the drawn dimensions by $\theta^{[\mathcal{D}']}$, and refer to the remaining parameters by $\theta^{[\mathcal{D} \backslash \mathcal{D}']}$. Similarly, we denote the search space for the drawn dimensions by $\Theta^{[\mathcal{D}']}$ and the best-observed values by $\theta^{*}=(\theta^{*})^{[\mathcal{D}']}\cup (\theta^{*})^{[\mathcal{D} \backslash \mathcal{D}']}$. We apply a Gaussian process surrogate model and optimize the acquisition function considering only the drawn dimensions (l. 7) as follows

\begin{align}
    \theta^{[\mathcal{D}']}_{\iota} \gets \arg max_{\theta^{[\mathcal{D}']}\in\Theta^{[\mathcal{D}']}} \mathcal{A}(\theta^{[\mathcal{D}']}).
\end{align}

We select the values for the remaining parameters, $\theta^{[\mathcal{D} \backslash \mathcal{D}']}_{\iota}$, by applying a \textit{Dropout-Mix} strategy \citep{li2018high}: With a probability of $p$, we randomly draw the values for the remaining parameters from their respective domains. Otherwise, with a probability of $(1-p)$, the parameter values yielding the best observed function value are copied for the left-out dimensions, i.e., $\theta^{[D \backslash D']}_{\iota}=(\theta^{*})^{[\mathcal{D}\backslash \mathcal{D}']}$ (ll. 8-13). We evaluate the target function and update the set of observations $R$ and incumbent parameters (ll. 14-16). Finally, we return the incumbent parameter setting $\theta^*$ (l. 18). 

\paragraph{B. Hierarchical Bayesian Optimization} We further present a hierarchical approach to apply BO in high-dimensional problem settings by decomposing the search space. Here, we optimize sets of hyperparameters sequentially, enabling the application of distinct surrogate models for each set of hyperparameters. 
Further, we decompose high-dimensional search spaces which exceed the capabilities of basic BO approaches, such that the resulting sub-problems can be solved via BO methods for low-dimensional search spaces. We show the pseudo-code for the hierarchical approach in Algorithm \ref{alg:HBO}.
\input{Algorithms/alg3}

As a basis, we assign the search space dimensions to disjoint sets $k=1,...,K$. We express the assignment of search dimension $\Lambda_j$ to set $k$ with a binary variable $\gamma_{\Lambda_jk}$:
\begin{equation}
  \gamma_{\Lambda_jk} =
    \begin{cases}
        1 & \text{if $\Lambda_j$ is assigned to set k}\\
        0 & \text{otherwise}
    \end{cases} \ \ \forall k=1,...,K, \Lambda_j \in \Theta, j = 1,...,J.
\end{equation}

 We assign each search dimension $\Lambda_j$ to exactly one set such that 
\begin{equation}
\sum_{k=1}^{K}{y_{{\Lambda_j}k}=1 \ \ \forall \Lambda_j \in \Theta, j = 1,...,J}.
\label{eq:disjointsets}
\end{equation}

We initialize the hyperparameters by applying random search (ll. 3-5). Second, we apply BO for all hyperparameter sets $k=1,...,K$ sequentially. Thus, we determine the search space $\Theta^{[k]}$ of set $k$ and refer to the parameters optimized in the set $k$ by $\theta^{[k]}$ (l. 7). We fix the remaining parameters denoted as $(\theta^*)^{[K\backslash k]}$ by copying the values of the incumbent parameter setting $\theta^{*}$ (l. 8). Before conducting BO for set $k$, we determine $m^{[k]}$ initial points via random search (l. 9). Before each function evaluation, we merge the varied parameters' values $\theta^{[k]}$ with the fixed parameters' values $(\theta^*)^{[K\backslash k]}$. We conduct $n^{[k]}$ iterations of BO for set $k$ (ll. 11-15). After optimizing the hyperparameters of set $k$, we continue with the next set until all sets of parameters have been optimized. 

%% file: Algorithms/alg1.tex
\begin{algorithm}
\caption{Bayesian Optimization}
\label{alg:BO}
\begin{spacing}{0.9}
\begin{algorithmic}[1]
\State Given: target function $f(\cdot)$, domain $\Theta$, number of iterations $n$, number of initial points $m$, acquisition function $\mathcal{A}$, surrogate model $\mathcal{M}$, observed results $R = \emptyset$
\State Output: Incumbent parameter setting  $\theta^{*}$

\State $R=\{(\theta_1, f(\theta_1)),...,(\theta_m, f(\theta_m))\} \gets \text{RandomSearch}(f(\cdot), \Theta, m)$

\State $\iota \gets m$
\For{$\iota \gets \iota+1 \ \text{to} \ \iota+n$}
    \State $\theta_{\iota} \gets \arg max_{\theta\in\Theta} \mathcal{A}(\theta)$
    \State $R \gets R \cup (\theta_{\iota},f(\theta_{\iota}))$

\EndFor

\State \textbf{return} $\theta^{*} = \arg \min_{\iota' \leq \iota} f(\theta_{\iota'})$
\end{algorithmic}
\end{spacing}
\end{algorithm}

%% file: Algorithms/alg2.tex
\begin{algorithm}
\caption{Bayesian Optimization with Dimension Dropout}\label{alg:BO_dropout}
\begin{spacing}{0.9}
\begin{algorithmic}[1]
\State Given: target function $f(\cdot)$, domain $\Theta$, number of iterations $n$, number of initial points $m$, 
acquisition function $\mathcal{A}$, observed results $R = \emptyset$, search dimensions $\mathcal{D}$, percentage share of search dimensions to be drawn $\Tilde{d}$, probability for fill-up strategy $p$
\State Output: Incumbent parameter setting  $\theta^{*}$
\State $R=\{(\theta_1, f(\theta_1)),...,(\theta_m, f(\theta_m))\} \gets \text{RandomSearch}(f(\cdot), \Theta, m)$
\State $\iota \gets m$
\For{$\iota=\iota+1 \ \text{to} \ \iota+n$}
    \State Randomly select a subset of dimensions $\mathcal{D}'$ such that $|\mathcal{D}'|=\left\lfloor\Tilde{d}*|\mathcal{D}|\right\rfloor$
    \State $\theta^{[\mathcal{D}']}_{\iota} \leftarrow \arg max_{\theta^{[\mathcal{D}']}\in\Theta^{[\mathcal{D}']}}\mathcal{A}(\theta^{[\mathcal{D}']})$
    \State $q \gets$ random number between 0 and 1
    \If {$q < p$}
        \State $\theta^{[\mathcal{D}\backslash \mathcal{D}']}_{\iota} \gets$ random values within domain $\Theta^{[\mathcal{D}\backslash \mathcal{D}']}$
    \Else
        \State $\theta^{[\mathcal{D}\backslash \mathcal{D}']}_{\iota} = (\theta^{*})^{[\mathcal{D}\backslash \mathcal{D}']}$
    \EndIf
    \State $\theta_{\iota} \leftarrow \theta^{[\mathcal{D}']}_{\iota} \cup \theta^{[\mathcal{D}\backslash \mathcal{D}']}_{\iota}$
    \State $R = R \cup {(\theta_{\iota}, f(\theta_{\iota})})$
    \State $\theta^{*} = \arg \min_{\iota' \leq \iota} f(\theta_{\iota'})$
\EndFor
\State \textbf{return} $\theta^{*}$
\end{algorithmic}
\end{spacing}
\end{algorithm}

%% file: Algorithms/alg3.tex
\begin{algorithm}

\caption{Hierarchical Bayesian Optimization}\label{alg:HBO}
\begin{spacing}{0.9}
\begin{algorithmic}[1]
\State Given: hyperparameter sets $k=1,...,K$, target function $f(\cdot)$, domain $\Theta$, domain of set $k$ $\Theta^{[k]}$, number of iterations per set $k$ $n^{[k]}$, number of initial points $m$, number of initial points for set $k$ $m^{[k]}$, acquisition function for set $k$ $\mathcal{A}^{[k]}$, 
observed results for set $k$ $R^{[k]} = \emptyset$
\State Output: Incumbent parameters  $\theta^{*}$

\State $R=\{(\theta_1, f(\theta_1)),...,(\theta_m, f(\theta_m))\} \gets \text{RandomSearch}(f(\cdot), \Theta, m)$
\State $\iota \gets m$
\State $\theta^{*} = \arg \max_{\iota' \leq \iota} f(\theta_{\iota'})$
\For{$k=1 \ \text{to} \ K$}
    \State $\theta^{[k]} \in \ \Theta^{[k]} = \{\Lambda_j | \Lambda_j \in \Theta\ \land \gamma_{\Lambda_jk}=1\}$ 
    \State $(\theta^*)^{[K\backslash k]} = \{ \lambda_j | \lambda_j \in \theta^* \land \gamma_{\Lambda_jk}=0\}$
    \State $R^{[k]} \gets \text{RandomSearch}(f(\cdot), (\theta^*)^{[K\backslash k]}, \Theta^{[k]}, m^{[k]})$
    \State $R \gets R \cup R^{[k]}$
    \For{$\iota = \iota+1 \ \text{to} \ \iota + n^{[k]}$}
        \State $\theta^{[k]}_{\iota} \gets \arg max_{\theta^{[k]}\in\Theta^{[k]}} \mathcal{A}^{[k]}(\theta^{[k]})$
        
        \State $\theta_{\iota} \gets \theta^{[k]}_{\iota} \cup (\theta^*)^{[K\backslash k]}$
        
        \State $R^{[k]} \gets R^{[k]} \ \cup \ (\theta_{\iota},f(\theta_{\iota}))$
    \EndFor
    \State $R \gets R \cup R^{[k]}$
    \State $\theta^{*} = \arg \min_{\iota' \leq \iota} f(\theta_{\iota'})$
\EndFor

\State \textbf{return} $\theta^{*}$

\end{algorithmic}
\end{spacing}
\end{algorithm}

%% file: contents/casestudy.tex
\section{Case Study: Ambulance Demand Prediction for Seattle}\label{sec:castestudy}

We conduct a numerical case study for Seattle's 911 call data\footnote{\label{fn:govdata}https://data.seattle.gov/} considering incidents of the category {\it{Medic Response}}, i.e., incidents requiring paramedical staff qualified for Advanced Life Support during 2 years (2020-2021). We divide this period into 8-hour intervals, corresponding to commonly applied 8-hour shifts in EMSs. By including ambulance demand of previous periods, we consider local, short-term dynamics such as local virus outbreaks during the Covid-19 pandemic. We divide Seattle into a grid of $11 \times 6$ subregions each with a dimension of approximately $2.5 \times 2.5 \texttt{ km}$ and apply a look back $L$ of 6 periods, i.e., we take into account the ambulance demand of two preceding days.

\paragraph{Features} We base our feature choice on the literature review in Section \ref{sec:literature} and focus on features that can only hardly be learned with our plain CNN architecture. We neglect infrastructural and demographic information, e.g., the mobility infrastructure or the age distribution among the population, as we only conduct short-term forecasts and these variables only change gradually over time. 

Since the weather changes dynamically, we include the following data from \textit{Weather Underground}\footnote{https://www.wunderground.com}: temperature, wind speed, humidity, dew point, sea level pressure, and precipitation. Except for the precipitation, we consider the daily minimum, maximum, and average for all weather features. As some of these inputs may be correlated, we conduct a correlation analysis and eliminate features with a correlation of $>80\%$. We present the results of the correlation analysis in Appendix \ref{sec:appendix_correlation_analysis}. We assume that the spatial distribution of weather data is negligible for our case study since the area is sufficiently small to assume that the measurements of different weather stations within this area are highly correlated. For larger areas, our generic approach enables the inclusion of two- or three-dimensional weather data, e.g., for (spatial) weather predictions or historic (spatio-temporal) weather data, correspondingly. 
 
However, \cite{wong2020effects} show that temporal temperature changes influence people's health conditions. Also, age, income, sex, and health conditions influence people's sensitivity to weather conditions. Thus, the impact of temporal weather changes may vary between subregions. Therefore, we perform transposed convolution on the historic one-dimensional weather data to learn its spatial impacts while the distribution of relevant socio-economic factors can be implicitly learned by the model.

Many studies include special-day effects for periods with an anomalous ambulance demand, e.g., during school holidays \citep{vile2012predicting} or New Year's Day \citep{channouf2007application}. To consider such effects, we take public holidays, school holidays, and events\footnote{https://data.seattle.gov/} into account. We represent holidays as binary features. As events take place at different locations across the depicted area, we include their locations on the two-dimensional, spatial level. We additionally consider the number of expected participants at each event to account for the events' sizes. 

We further include time information, as \cite{channouf2007application} show that the day-of-week, month-of-year, and hour-of-day impact ambulance demand. We embed this information as one-hot-encoded vectors. To increase the models' performance, we scale the input data such that all values range within $[0,1]$.

\paragraph{Benchmarks} The developed CNN serves as a representative example of an integrated spatio-temporal ANN architecture. To evaluate the numerical benefits of incorporating spatio-temporal dependencies in such architectures, we compare this approach to an iterative ANN architecture, applying an MLP as an exemplary model. MLPs are fully connected feedforward ANNs commonly applied for forecasting tasks, also in the domain of ambulance demand prediction \citep{setzler2009ems,chen2015demand}. We refer to Equation \ref{eq:denselayer} for a definition of a fully connected architecture. We further aim at evaluating the performance of the proposed intrinsic feature selection algorithm which integrates feature selection in the hyperparameter tuning process of the corresponding ANN architectures. For this evaluation, we compare our approach to other commonly applied intrinsic feature selection methods that are able to solve prediction tasks. One approach widely applied for regression tasks and capable of providing valuable insights into features' importance which make outcomes interpretable for practitioners are decision trees. However, decision trees are prone to overfitting, especially for high dimensional search spaces as there is mostly an insufficient number of samples for each parameter value. To apply a model more robust to overfitting, we additionally apply random forests, i.e., ensembles of multiple decision trees based on randomly selected data samples. The average prediction generated by the individual decision trees constitutes the random forest's prediction. In line with \cite{setzler2009ems}, \cite{zhou2015spatio} and \cite{zhou2016predicting}, we further compare the performance of the developed CNN to a common industry practice, i.e., the Medic method. This method takes the average ambulance demand of similar time periods of the past 4 weeks, i.e., the historic ambulance demand associated with the same weekday and time, of the current and preceding years as prediction. To enable a fair comparison to the ANNs trained on data from 2020-2021, we consider the same data range for the Medic method.

\paragraph{Hyperparameter Tuning} As the hyperparameter choices for decision trees and random forests are limited, we apply grid search for tuning. However, for the CNN and MLP, we optimize the hyperparameters by applying i) random search ii) BO with Gaussian process, random forest, and extremely randomized tree surrogate models iii) BO with dimension dropout, and iv) the novel hierarchical BO with Gaussian process, random forest, and extremely randomized tree surrogate models. First, we perform these approaches without feature selection, and second, including feature selection. We present the hyperparameters and their domains in Appendix \ref{sec:appendix_hyperparameters}. We further show the set assignments for the hierarchical BO. 

We determine the order in which we optimize the hyperparameter sets based on two criteria: The degree to which preliminary experiments can yield adequate values, and, the size of the search space. 
The better we can determine adequate hyperparameter values, the later we optimize them. This ensures that we optimize the hyperparameters that are more complex to determine on an appropriate basis, i.e., the hyperparameters that are currently not optimized are set to well-performing values which were either determined by applying random search for initialization, or, optimized at a previous level. Later, we can fine-tune the hyperparameters we fixed in the first place. 

For our application, we distinguish three levels. At each level, one parameter set is tuned. The first level focuses on the main architecture decisions, such as the number of layers and filters per layer. The second level tunes the activation functions. The third level controls the regularization mechanisms to avoid overfitting and decides on the batch size, learning rate, and optimizer. In the case of incorporating feature selection, we include an additional binary hyperparameter for each feature, stating whether the feature is included (1) or not (0). In the hierarchical BO, we therefore include an additional level for these decisions. We optimize this level first, such that we base the choice of the remaining hyperparameters tuning the model on the derived features. 

\paragraph{Training} We train our model on 60\% of the data and split the remaining data into a test and a validation set, each containing 20\% of the data. We apply a mean-squared-error (MSE) loss function and train the model for 300 epochs. In the case that the validation loss does not improve for 20 epochs, we abort the training process and measure the MSE on the test set. We initialize the hyperparameters of the hierarchical BO based on 500 iterations of random search. To enable a comparison between the hierarchical approach and the remaining methods, we initialize them equally. After initialization, we execute 1000 iterations. As the number of parameters optimized at each level of the hierarchical BO is limited, we reduce the number of iterations to 250, applying random search for the first 25 iterations per hyperparameter set. 

We implement the algorithms in Python 3.8 using TensorFlow 2.10.1. We perform BO using Scikit-Optimize 0.9.0, which we adapt to apply dimension dropout. We train the CNNs on an NVIDIA A100 GPU with 80 GB RAM. The MLPs are trained with an Intel(R) Xeon(R) processor E5-2697 v3 with 15.7 GB RAM.

%% file: contents/results.tex
\section{Experimental Results}\label{sec:experimentalresults}

We first compare the performance of the developed CNN architecture against its benchmarks and show the results we obtain for the different hyperparameter tuning approaches. Second, we analyze the selected features when conducting feature selection and calculate the contribution of each feature to the prediction. Third, we investigate the CNN's performance for different forecasting time intervals.

\subsection{Model and Hyperparameter Tuning Performances}\label{sec:experimentalresults_modelperformances}

In Table \ref{table:results} we compare the MSE of the developed CNN architecture and its benchmark models, i.e., the Medic method, MLPs, decision trees, and random forests. The learning-based approaches were tuned with the introduced hyperparameter optimization approaches and initialized equally with random search.

\input{Tables/Results}

As can be seen, the incumbent CNN achieves a 9.83\% lower MSE than the best MLP. The CNN further outperforms the Medic method, decision tree, and random forest by 9.98\%, 14.84\%, and 11.26\%. As expected, we see that the decision tree badly generalizes to unseen data leading to an incumbent tree with a depth of five. The incumbent random forest has a depth of eight. We visualize their tuning results in Appendix \ref{sec:appendix_hyperparameters}. Although the random forest performs better on new data, it still remains inferior to the ANNs. Both tree-based approaches are unable to identify relevant patterns due to their limited depth and the resulting negligence of relationships among external features. Although the Medic method bases its prediction solely on historic ambulance demand neglecting all external features, it still outperforms the tree-based approaches and yields comparable results to the MLP. Contrarily, the CNN benefits from its ability to detect spatio-temporal patterns across subregions within historic ambulance demand and combines these patterns with external information. This ability also encourages the use of the presented feature selection approach which enables the application of the developed CNN instead of the intrinsic benchmarks, i.e., decision trees, and random forests. The advantage of the introduced intrinsic approach is the exchangeability of the model: We are not bound to an intrinsic tree-based approach to make the prediction. Instead, we can apply the developed CNN resulting in the lowest MSE. We further observe that retaining less than 50\% of the features in the CNN reduces the number of trainable parameters by 40.4\% and only slightly increases the MSE by less than 0.05\%. For practitioners, this performance decrease remains negligible and a leaner model is easier to maintain, as less input data must be collected, validated, and pre-processed.

{\result{The CNN architecture outperforms the incumbent MLP, Medic method, random forest, and decision tree by 9.83\%, 9.98\%, 11.26\%, and 14.84\%, correspondingly.}}\vspace*{0.5\bigskipamount}
{\result{The tree-based approaches badly generalize to unseen data, resulting in shallow trees which are unable to identify underlying patterns.}}\vspace*{0.5\bigskipamount}
 {\result{Incorporating feature selection in the hyperparameter tuning process reduces the number of trainable parameters by 40.4\% while the performance decrease of less than 0.05\% remains negligible.}}\vspace*{0.5\bigskipamount}

\begin{figure}[ht]
\centering
  \includegraphics[width=.75\linewidth]{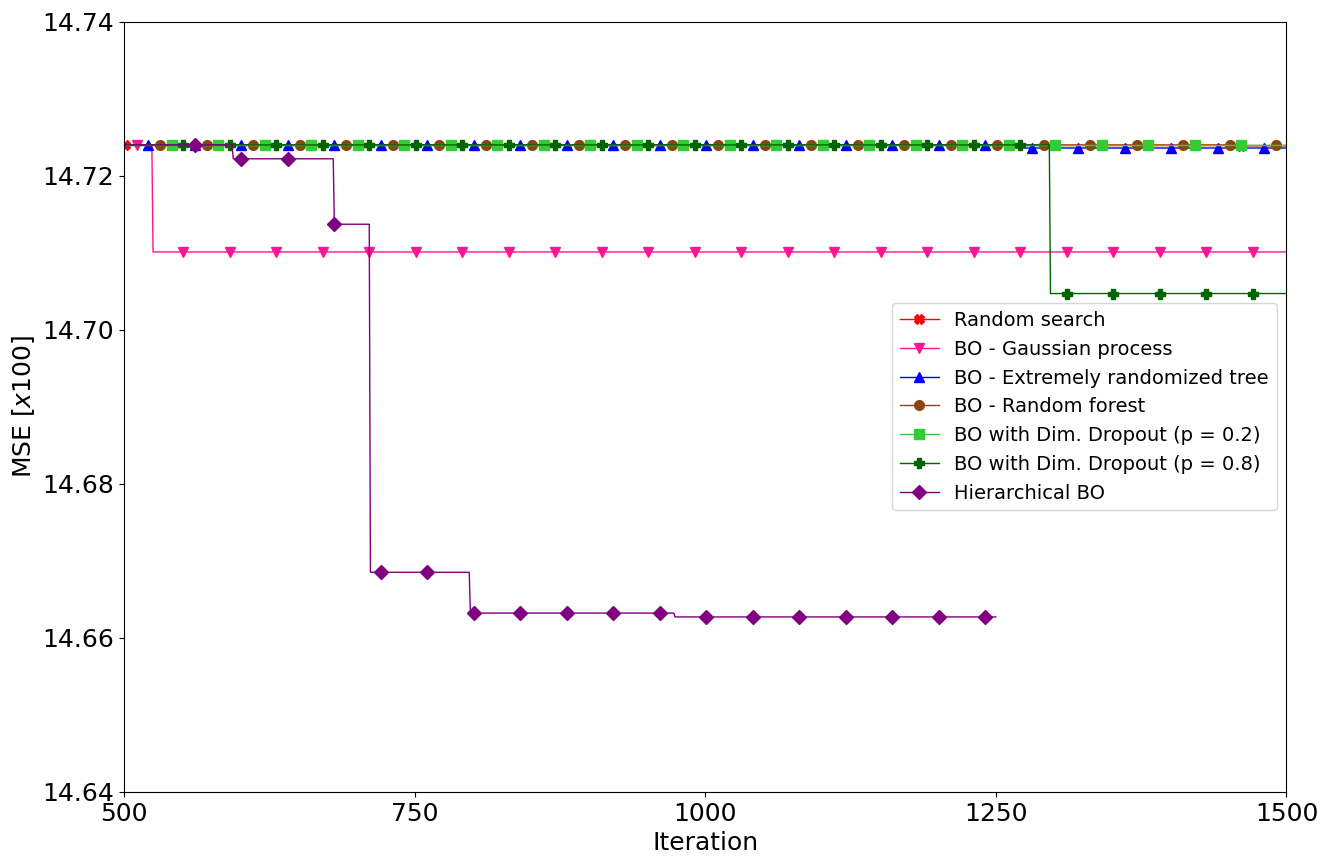}
  \caption{\centering{Convergence of the hyperparameter tuning approaches of the CNN without Feature Selection (after applying 500 iterations random search for initialization)}}
  \label{fig:convergence_cnn}
\end{figure}

\begin{figure}[ht]
\centering
  \includegraphics[width=.75\linewidth]{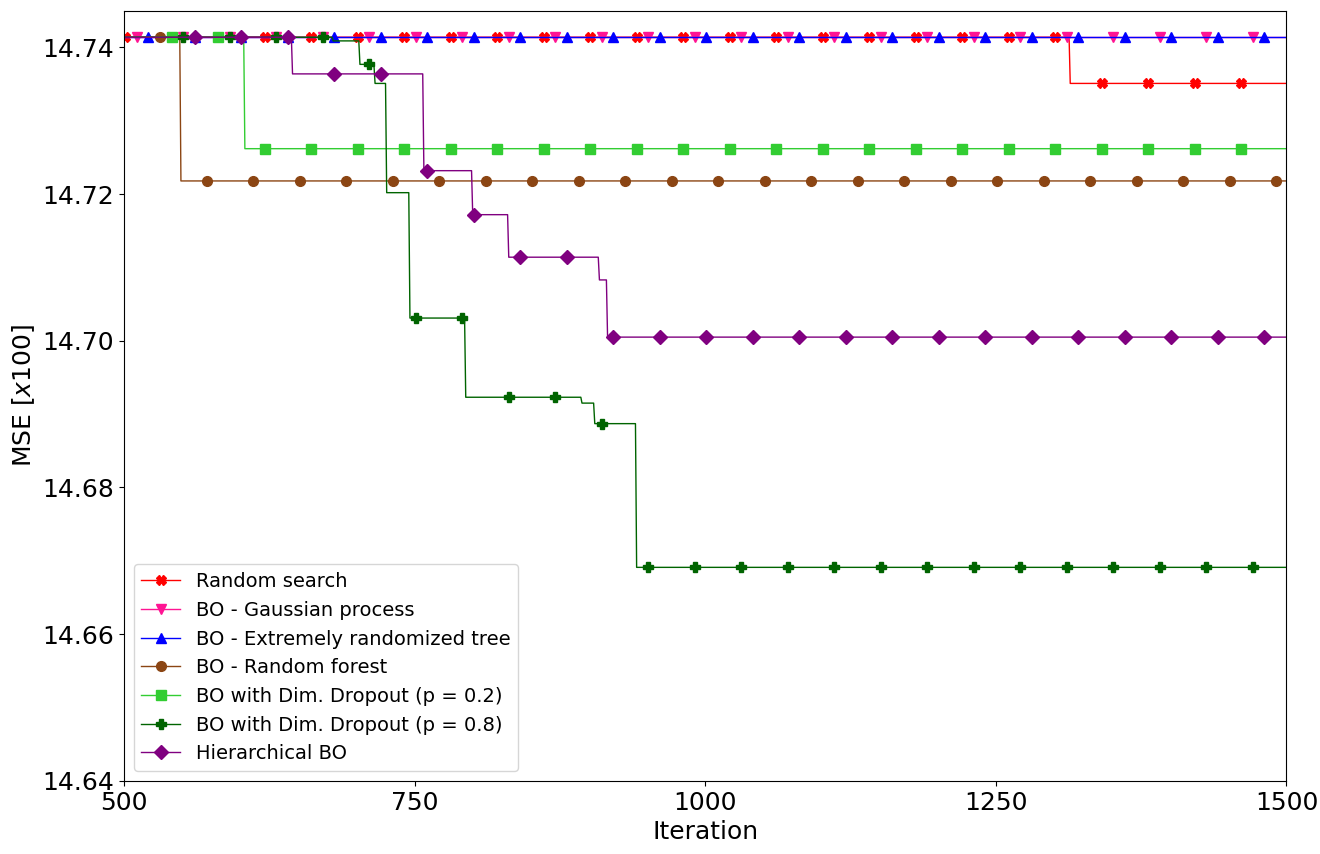}
  \caption{\centering{Convergence of the hyperparameter tuning approaches of the CNN with Feature Selection (after applying 500 iterations random search for initialization)}}
  \label{fig:convergence_cnn_2}
\end{figure}

When including feature selection decisions in the hyperparameter tuning process, we integrate a binary variable for each feature selection decision. This exponentially increases the search space. However, it enables the model to simultaneously decide on features and make hyperparameter decisions. This is beneficial, as the CNN's architecture, e.g., the number of layers in the model, depends on the selected features. Therefore, we evaluate the influence of including feature selection in the hyperparameter tuning process. We compare the convergence of the different tuning approaches with and without feature selection for the CNN in Figures \ref{fig:convergence_cnn} and \ref{fig:convergence_cnn_2}. Based on the hyperparameters determined by the corresponding tuning approach in each iteration, we show the MSE achieved on the test set after training the model.
In Figure \ref{fig:convergence_cnn} we neglect feature selection in the hyperparameter tuning process. Here, the basic BO can only slightly decrease the MSE in 1000 iterations when utilizing a Gaussian process surrogate model. In contrast, BO with dimension dropout and the hierarchical BO can better improve the MSE. In Figure \ref{fig:convergence_cnn_2} we include feature selection decisions. Here, the basic BO can only slightly decrease the MSE in 1000 iterations when utilizing a random forest surrogate model. For including feature selection, we add a binary variable for each selection decision, which a tree-based surrogate can better optimize than Gaussian processes. However, for both cases, i.e., with and without feature selection, the BO with dimension dropout and the hierarchical BO are superior to tackle the high dimensional search space.

{\result{The BO with dimension dropout and the hierarchical BO outperform basic BO approaches and random search.}}\vspace*{0.5\bigskipamount}


\subsection{Features' Importance}

\begin{figure}[ht]
\centering
\includegraphics[width=0.9\textwidth]{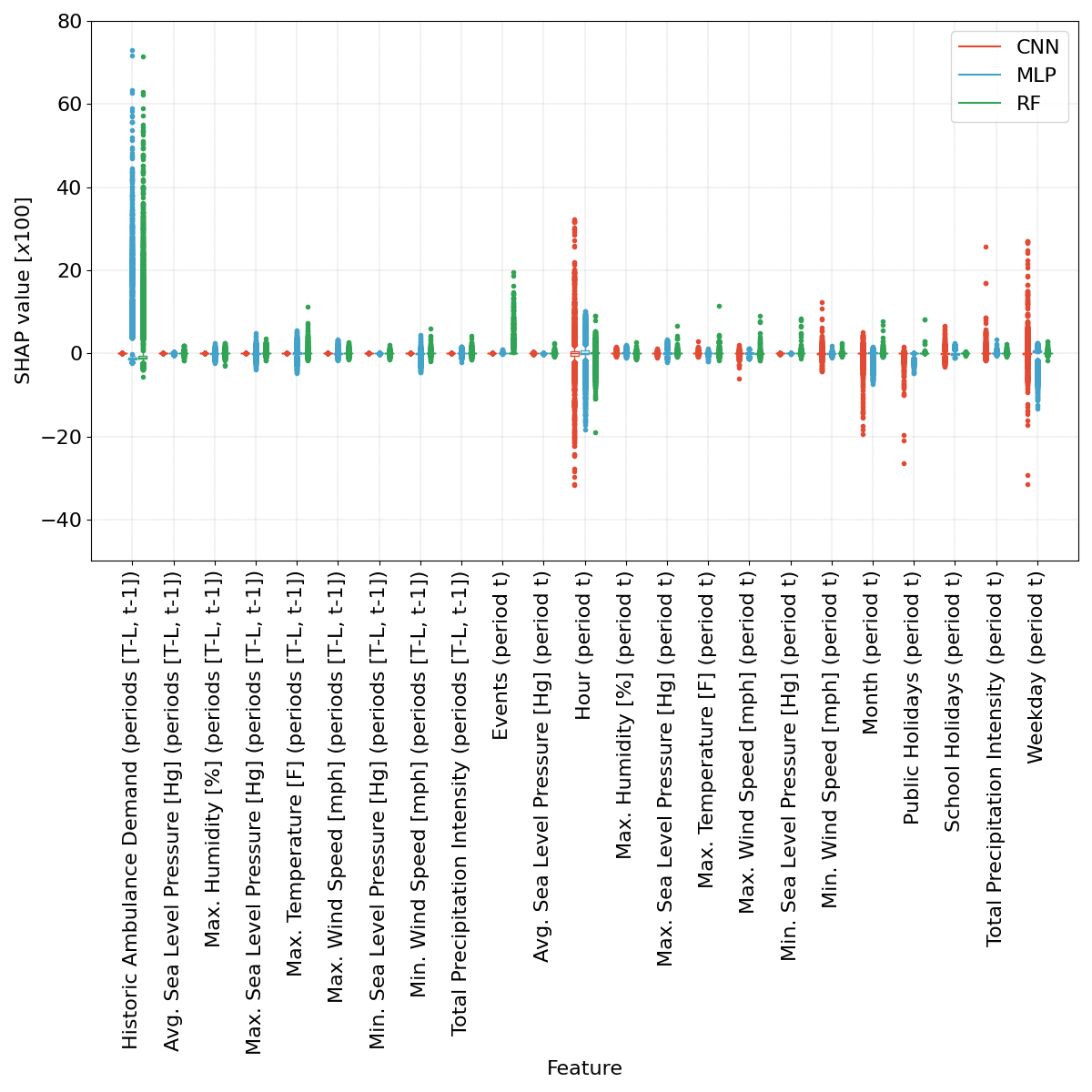}
{\caption {\centering Features' Importance: SHAP values} \label{fig:feature_importances}}
\end{figure}

To further investigate the reasons for the CNN's superiority, we analyze the importance of the external features for the different models. We base this analysis on the incumbent models derived in Section \ref{sec:experimentalresults_modelperformances} for 8-hour time-intervals.
Here, we distinguish two cases: first, when including feature selection in the hyperparameter tuning approach, we present the selected features in Table \ref{table:selected_features}. Second, when excluding feature selection in the hyperparameter tuning approach, we keep all features and present their importance in Figure \ref{fig:feature_importances}. Since the decision tree always conducts feature selection, we only present its selected features. As the random forest is an ensemble of decision trees with randomly selected features, we only present its features' importance.

\paragraph{Selected Features}

\input{Tables/selected_features}
Table \ref{table:selected_features} shows that the amount and type of features selected by the ANNs and decision tree strongly vary. The tuning process of the decision tree reveals that the tree overfits for high depths, i.e., we observe that the model's performance decreases on unseen data when increasing the tree's depth beyond five, visualized in Appendix \ref{sec:appendix_hyperparameters}. We further see that the decision nodes' conditions made at depths below four are exclusively based on historic ambulance demand. Only in the final splits, weather features, e.g., the maximum wind speed of past periods, or the time-of-day can be decisive. As these external features are only used by decision nodes preceding the leaf nodes, the model cannot identify relevant patterns among these features. 

{\result{The decision tree only uses few external features in decision nodes preceding the leaf nodes, preventing it from identifying patterns among these features.}\label{res:feature_selection_decision_tree}}\vspace*{0.5\bigskipamount} 

In contrast to the decision tree, the ANNs combine information from multiple external features. Table \ref{table:selected_features} shows that although the features selected by the CNN and MLP differ, they mostly include similar types of information; while one model selects a feature with predicted data, the other model includes the same type of information but includes the feature with the corresponding historic data, and vice versa. 
For example, both models include information about the sea level pressure and humidity. This is in line with many studies showing the effect of humidity and air pressure on people's health conditions influencing ambulance demand, e.g., the risk for cardiovascular diseases \citep{ou2014impact, borghei2020relationship}. In addition, both models include the hour and weekday which confirms their relevance observed in earlier studies \citep{setzler2009ems, matteson2011forecasting}. 
Simultaneously, some features are neglected by all models. As we include historic ambulance demand of previous periods, some features whose values remain unchanged for several periods, can be indirectly learned by the model. For this reason, the month and school holidays, mostly lasting several days, are neglected. 

{\result{The CNN and MLP select common types of information, such as weather data that has already been identified in medical studies to harm people's health conditions. In addition, in line with earlier studies, our results confirm the relevance of considering time information, i.e., the time-of-day and weekday.}\label{res:feature_selection_mlps}}\vspace*{0.5\bigskipamount}

\paragraph{Features' Importance} In Figure \ref{fig:feature_importances} we present the features' importance derived for the case of retaining all features during the tuning process. To express the importance, we calculate the SHAP values using shap 0.41.0 \citep{lundberg2017a}. We determine these SHAP values for 50 randomly selected time periods covering all subregions from the test set. Each value allows us to quantify the feature's contribution to the corresponding prediction compared to the base value. We calculate the average prediction output over 300 random time periods over all subregions from the training set as a base value.

 While historical data highly contributes to the prediction of the MLP and random forest, the CNN's output is mainly influenced by data characterizing the period to be predicted. This indicates that the CNN is more robust to changes in the upsampled and three-dimensional input data. Contrarily, the MLP and random forest are more sensitive to such changes since they conduct local forecasts for each subregion separately. Thus, changes in the historical data of the corresponding subregion strongly influence the forecast. In contrast, the CNN considers the historic data of all subregions simultaneously. Thus, it is more robust to changes in the historic data of a single subregion, which is advantageous, e.g., for handling demand outliers or data inconsistencies.
 
 {\result{The features' importance strongly differ between the MLP, random forest, and CNN. The CNN is more robust against changes in the upsampled data and historic ambulance demand.}}

\subsection{Sensitivity Analysis: Time Granularity}
\begin{figure}[ht]
\centering
\includegraphics[width=.75\textwidth]{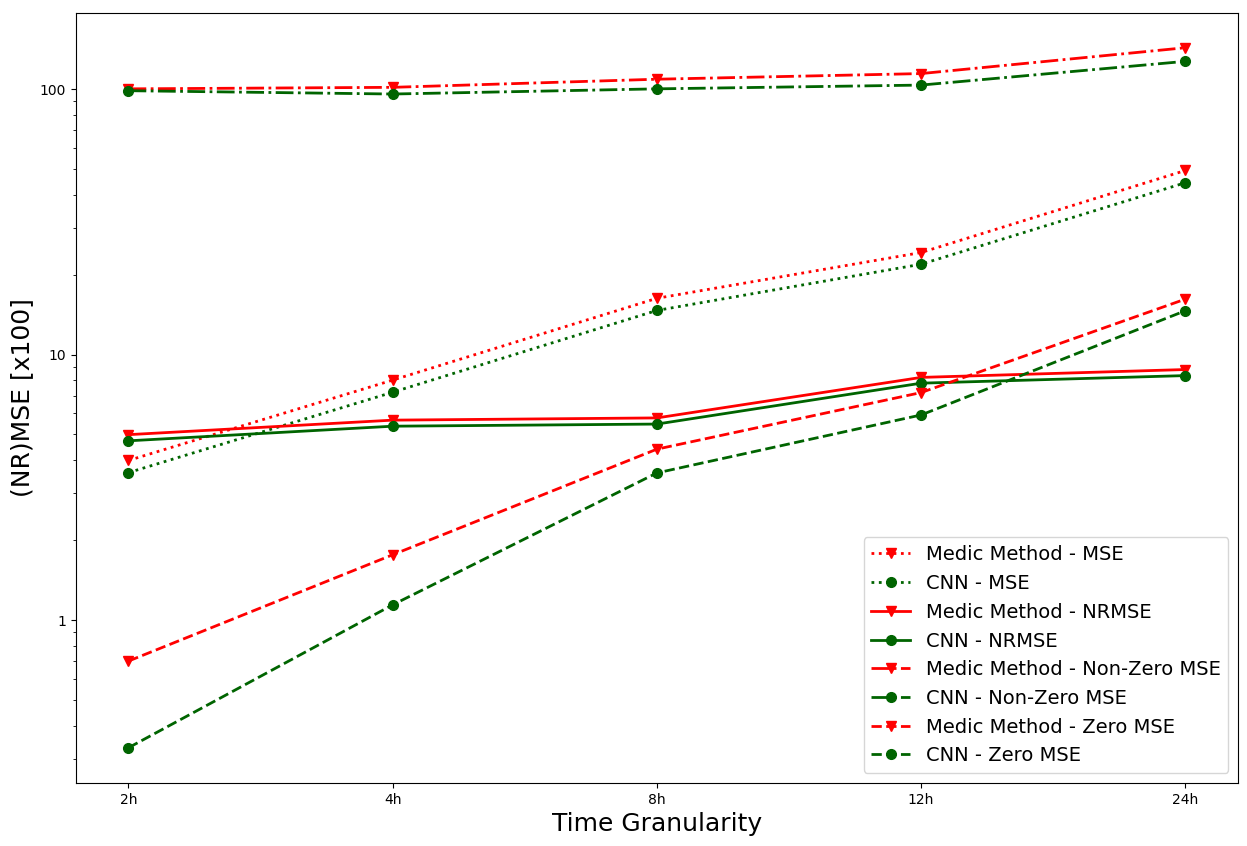}
{\caption {\centering Sensitivity Analysis: Time Granularity} \label{fig:sensitivity_analysis}}
\end{figure}

Earlier studies address the challenge arising from sparse ambulance demand which we face in the case of fine spatio-temporal granularities \citep{zhou2015spatio,zhou2016predicting}. For an extensive study of an iterative ANN architecture, we refer to \cite{setzler2009ems} who investigate the influence of fine spatio-temporal granularities on the performance of an MLP and compare it to the Medic method. To close this gap for the domain of integrated spatio-temporal ANN architectures, we conduct a sensitivity analysis and investigate the CNN's performance for five different time intervals: 2 hours, 4 hours, 8 hours, 12 hours, and 24 hours. Similar to \cite{setzler2009ems}, 
we compare the performance of the developed CNN against the Medic method. Previous results show that the hierarchical BO and BO with dimension dropout yield superior results for high-dimensional search spaces. For this reason, we apply these approaches in the following experiments when tuning the CNN. For both models, we calculate the MSE obtained on the test set for each time interval. In addition, we compute the MSE for instances with zero and non-zero demands individually. Moreover, to take the amplitude of each time interval into account, we include the normalized root-mean-square error (NRMSE) in the analysis, calculated as follows

\begin{equation}
    NRMSE = \frac{\sqrt{MSE}}{y_{max} - y_{min}},
\end{equation}

where $y_{max}$ and $y_{min}$ are the maximum and minimum observed ambulance demand across all time intervals and subregions. We show the obtained MSEs and NRMSEs for the incumbent CNN and Medic method in Figure \ref{fig:sensitivity_analysis}.

For all analyzed performance measures and time horizons, the CNN outperforms the Medic method. We see that both demand types, i.e., instances with zero and non-zero ambulance demand, can be better predicted by the CNN. The lower the granularity, the better the performance of the CNN for non-zero demands compared to the Medic method. This indicates, that in the case of higher demands, the CNN can better identify underlying patterns within the data. Considering the time interval's amplitude in this analysis, we further see that the NRMSE remains stable across all time intervals for both models. This shows that the prediction performance remains stable if we take into account the time intervals' emergency call volumes. Disregarding these amplitudes, the MSEs and the NRMSEs improve for finer time granularities due to the reduction in ambulance demand and the increase in zero-demands which can be well predicted. 

 {\result{The CNN outperforms the Medic method for all analyzed time intervals. Both demand types, i.e., instances with zero and non-zero ambulance demand, can be better predicted by the CNN.}}

%% file: Tables/Results.tex
\begin{table}[ht]
\centering
\begin{adjustbox}{width=\textwidth}

\begin{tabular}{ c || c | c | c  c | c  c  c | c c | c c c c || c}

&No tuning&Grid Search&\multicolumn{2}{c|}{Random Search}&\multicolumn{3}{c|}{BO}&\multicolumn{2}{c|}{\shortstack{BO with\\ Dim. Dropout}}&\multicolumn{4}{c ||}{Hierarchical BO}&\\\cline{2-14}

&&&&&\multicolumn{3}{c|}{surrogate model}&p=0.2&p=0.8&\multicolumn{4}{c||}{set}&Feat.\\

Model&&&&&GP&RF&ET& $\Tilde{d}$=0.25&$\Tilde{d}$=0.25&{{k=0}}&{{k=1}}&{{k=2}}&{{k=3}}&Sel.\\\hline

CNN&-&-&14.74&14.74&14.74&14.74&14.72&14.73&\textbf{14.67}&14.74&14.70&14.70&14.70&\checkmark\\
CNN&-&-&14.72&14.72&14.71&14.72&14.72&14.72&14.70&-&14.67&14.66&\textbf{14.66}&\\
MLP&-&-&16.34&16.31&16.34&16.30&16.28&16.27&\textbf{16.26}&16.27&16.27&16.27&16.27&\checkmark\\
MLP&-&-&16.32&16.32&16.32&16.29&16.28&16.28&16.29&-&16.29&16.29&\textbf{16.28}&\\

RF&-&\textbf{16.52}&-&-&-&-&-&-&-&-&-&-&-&\checkmark\\
DT&-&\textbf{17.22}&-&-&-&-&-&-&-&-&-&-&-&\checkmark\\
Medic&\textbf{16.29}&-&-&-&-&-&-&-&-&-&-&-&-&\checkmark\\

\hline
\multirow{2}{*}{\# iter.}&&&\multirow{2}{*}{500}&\multirow{2}{*}{1500*}&\multicolumn{5}{c |}{\multirow{2}{*}{1500*}}&&750*&1000*&1250*&\\

&&&&&\multicolumn{5}{c |}{}&750*&1000*&1250*&1500*&\checkmark\\\hline

\end{tabular}
\end{adjustbox}
\caption{{Comparison of MSE (*incl. 500 iterations random search for initialization) \textbf{$[\times 100]$.}}
\textmd{We compare the MSE of the convolutional neural networks (CNNs), multilayer perceptrons (MLPs), decision trees (DTs), random forests (RFs) and the Medic method. For hyperparameter tuning, we apply Bayesian optimization (BO) with Gaussian process (GP), RF and extremely randomized tree (ET) surrogate models. Hierarchical BO is applied for sets $k\in\{1,2,3\}$. We add set k=0 when conducting feature selection. We train the model on 60\% of the data and apply a validation and test set, each containing 20\% of the data.}}
\label{table:results}

\end{table}

%% file: Tables/selected_features.tex
\begin{table}[h!]
\centering
\begin{adjustbox}{width=.8\textwidth}

\begin{tabular}{ l || c c c c c c c c | c c c c c c c c c c c c c c}

Features&\multicolumn{8}{c|}{Historic data from periods $[t-L, t-1]$}&\multicolumn{14}{c}{Data for predicted period $t$}\\\hline
&{\rotatebox{90}{Avg Sea Level Pressure [Hg]}}&{\rotatebox{90}{Max. Humidity [\%]}}&{\rotatebox{90}{Max. Sea Level Pressure [Hg]}}&{\rotatebox{90}{Max. Temp [F]}}&{\rotatebox{90}{Max. Wind Speed [mph]}}&{\rotatebox{90}{Min. Sea Level Pressure [Hg]}}&{\rotatebox{90}{Min. Wind Speed [mph]}}&{\rotatebox{90}{Total Precipitation Intensity}}&{\rotatebox{90}{Events}}&{\rotatebox{90}{Avg. Sea Level Pressure [Hg]}}&{\rotatebox{90}{Hour}}&{\rotatebox{90}{Max. Humidity [\%]}}&{\rotatebox{90}{Max. Sea Level Pressure [Hg]}}&{\rotatebox{90}{Max. Temp [F]}}&{\rotatebox{90}{Max. Wind Speed [mph]}}&{\rotatebox{90}{Min. Sea Level Pressure [Hg]}}&{\rotatebox{90}{Min. Wind Speed [mph]}}&{\rotatebox{90}{Month}}&{\rotatebox{90}{Public Holiday}}&{\rotatebox{90}{School Holiday}}&{\rotatebox{90}{Total Precipitation Intensity}}&{\rotatebox{90}{Weekday}}\\\hline

CNN&\checkmark&\checkmark&\checkmark&-&-&-&-&-&-&\checkmark&\checkmark&\checkmark&-&-&\checkmark&\checkmark&-&-&\checkmark&-&-&\checkmark\\

MLP&\checkmark&-&-&-&\checkmark&\checkmark&-&\checkmark&-&\checkmark&\checkmark&\checkmark&\checkmark&-&-&-&-&-&\checkmark&-&\checkmark&\checkmark\\

Decision Tree&-&-&-&\checkmark&\checkmark&-&-&-&-&-&\checkmark&-&-&-&-&-&-&-&-&-&\checkmark&-\\

\end{tabular}
\end{adjustbox}
\caption{Selected Features}
\label{table:selected_features}

\end{table}

%% file: contents/conclusion.tex
\section{Conclusion}\label{sec:conclusion}

We presented a novel CNN architecture that transforms time series information into heatmaps to predict ambulance demand. It can incorporate historical ambulance demand and external features of varying dimensions and uses three-dimensional convolutional layers to detect correlations in space and time. We further introduced a hyperparameter tuning framework utilizing BO that enables to intrinsically tune the hyperparameters while selecting features. To tackle the high-dimensional search space, we applied BO with dimension dropout and introduced a novel hierarchical BO. We further calculated SHAP values to investigate the contribution of each feature to the prediction. We further analyzed the CNN's performance for different forecasting horizons. To apply the developed CNN to real data, we conducted a numerical case study for Seattle's 911 call data.

Results show that the CNN outperforms all benchmarks by more than 9\%. Although the ANNs select different features, they include similar types of information confirming their relevance observed also in earlier studies in the domain of ambulance demand prediction and medical studies investigating the influence of weather data on people's health conditions. We further showed that the CNN is more robust against changes in the three-dimensional and upsampled input data. When including feature selection in the hyperparameter tuning process, we observed that excluding more than 50\% of the features reduces the trainable parameters by 40.4\% while the performance loss of $<0.05\%$ is negligible. Moreover, the CNN is superior to the current industry practice for all analyzed time horizons, ranging between 2-hour and 24-hour time-intervals.

In future work, the CNN architecture could be further optimized for fine space and time granularities, e.g., by giving non-zero demands more weight. Moreover, the derived features' importance could be used to improve the feature selection process. In addition, further investigating the convergence behavior of the BO when including feature selection decisions remains an interesting subject for future research.

%% file: contents/appendix.tex
\section{Appendix}\label{sec:appendix}

\subsection{CNN Architecture}\label{sec:cnn_architecture}
 Table \ref{table:cnn_architecture} presents the CNN architecture. $X^{\text{1D}}_{i}=\{{X^{\text{1D}}_{i}}^{\text{T}}, {X^{\text{1D}}_{i}}^{\Bar{\text{T}}}\}$, $X^{\text{2D}}_{i}$, $X^{\text{3D}}_{i}$, and $X^{\text{S}}_{i}$ represent the sets of one-dimensional, two-dimensional, three-dimensional, and scalar inputs for data instance $i$, where ${X^{\text{1D}}_{i}}^{\text{T}}$ denotes the set of one-dimensional data which is upsampled and ${X^{\text{1D}}_{i}}^{\Bar{\text{T}}}$ refers to one-dimensional data that is not upsampled. We denote the length of input vector $x_{i_j} \in {X^{\text{1D}}_{i}}^{\Bar{\text{T}}}$ by $l_{j}$.
\input{Tables/cnn_architecture}

\subsection{Correlation Analysis for Weather Data}\label{sec:appendix_correlation_analysis}
\begin{figure}[h!]
  \includegraphics[width=.9\linewidth]{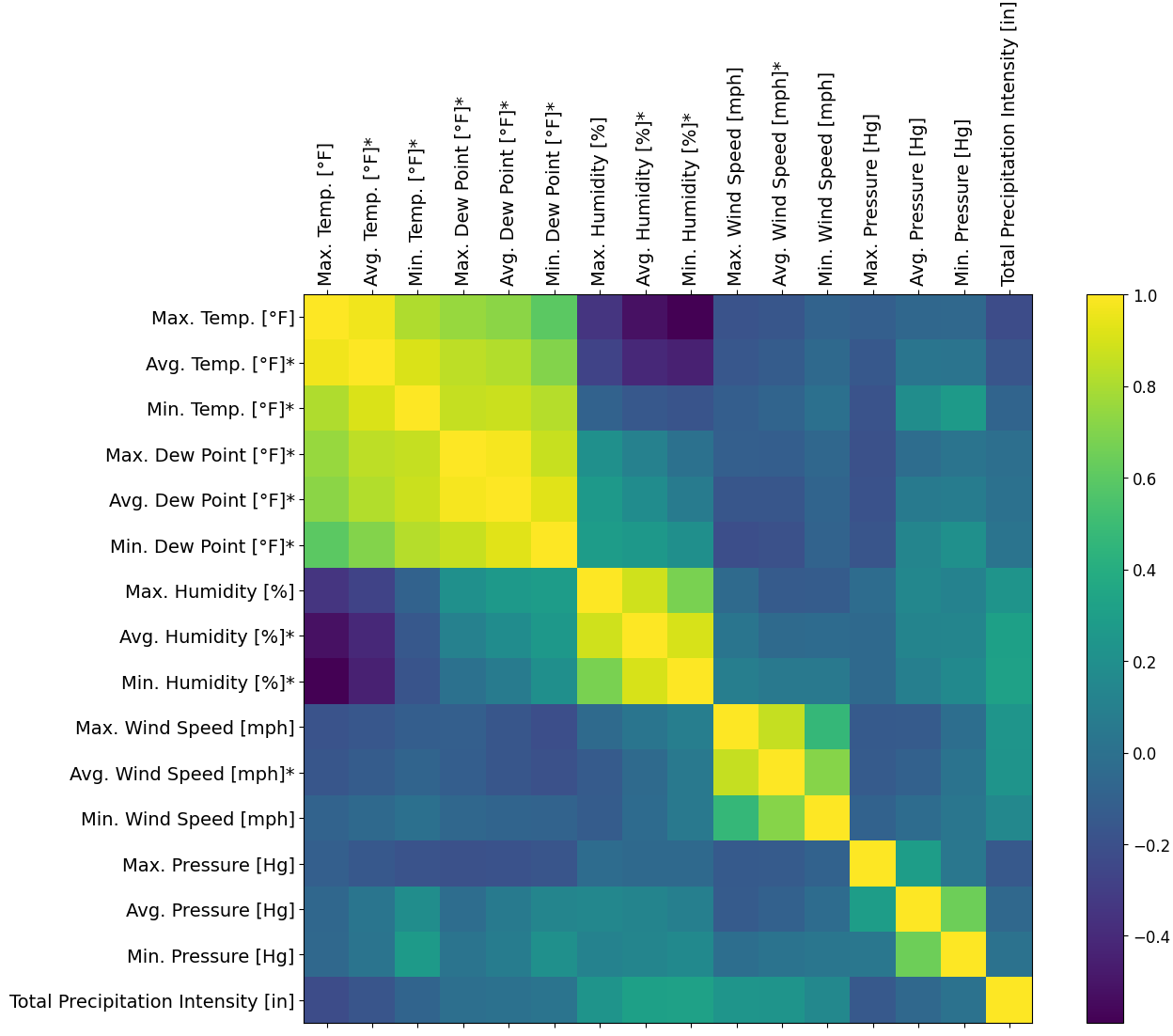}
  \caption{Correlation Matrix of Weather Features (* excluded due to a correlation of $>80\%$)}
  \label{fig:correlation}
\end{figure}

In Figure \ref{fig:correlation} we show the correlations between the weather features included in our data set. We preserve only one feature among those with a correlation greater than 80\%: We exclude the average wind speed, the average, and minimum temperature, the average, and minimum humidity, as well as the maximum, average, and minimum dew point.

\subsection{Hyperparameter Domains \& Tuning Results}\label{sec:appendix_hyperparameters}
\begin{figure}
\centering
  \includegraphics[width=.7\linewidth]{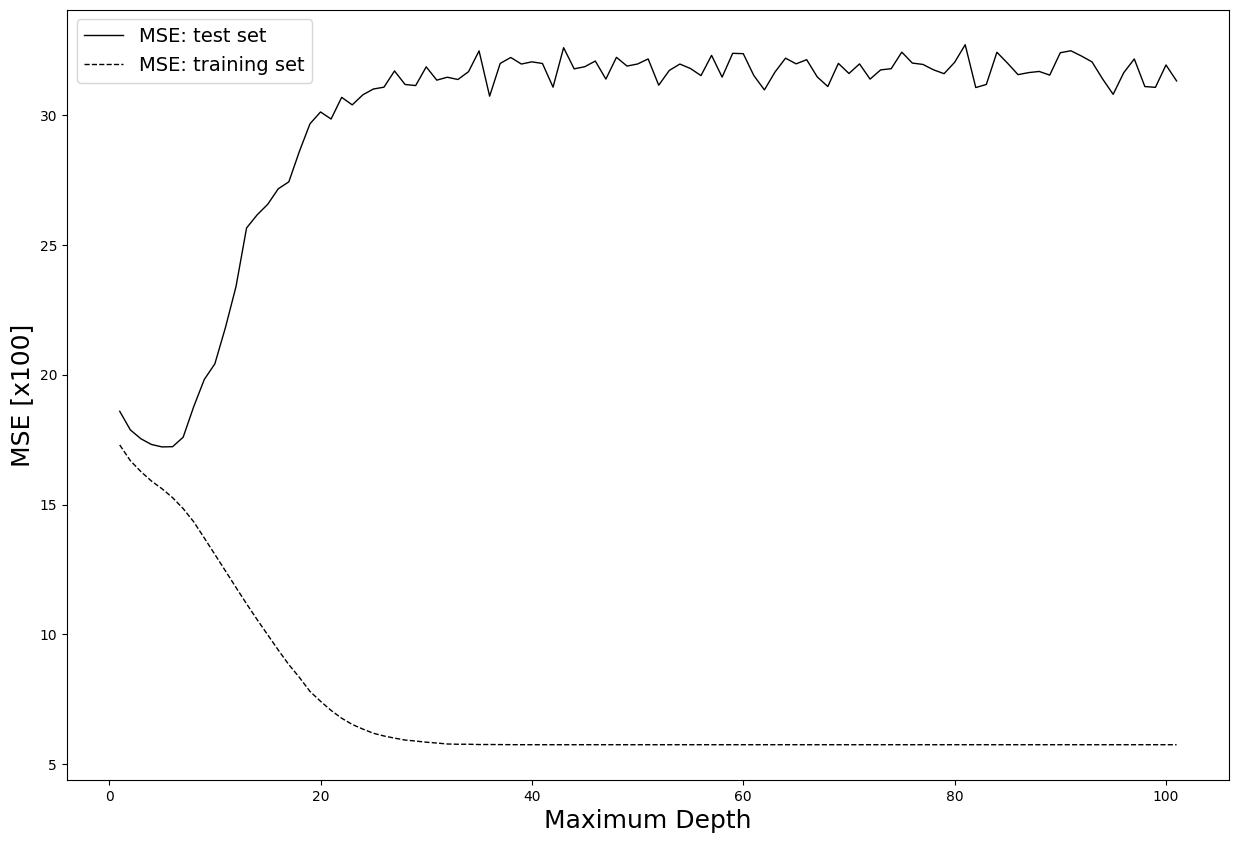}
  \caption{Decision Tree Hyperparameter Tuning}
  \label{fig:dt_tuning}
\end{figure}

\begin{figure}
\centering
  \includegraphics[width=.8\linewidth]{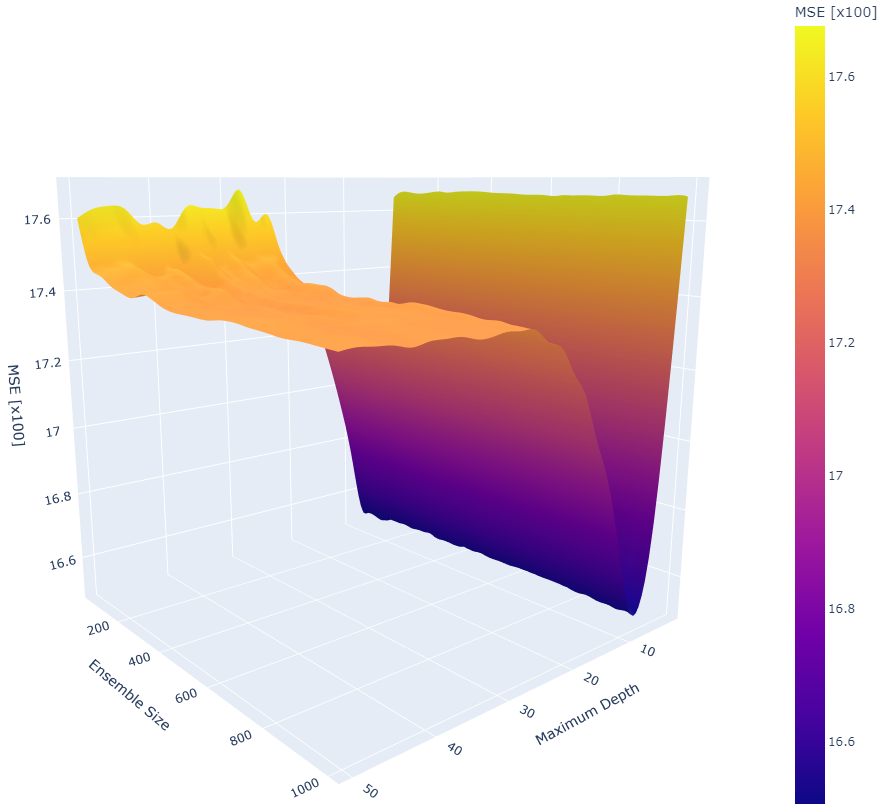}
  \caption{Random Forest Hyperparameter Tuning}
  \label{fig:rf_tuning}
\end{figure}

In Tables \ref{table:CNN_Search_Space} - \ref{table:GS_Trees} we show the hyperparameter domains for the CNN, MLP, random forest, and decision tree. We further present the set assignments considered for the hierarchical BO applied for the CNN and MLP. We show the results of different tuning approaches of the ANNs in Section \ref{sec:experimentalresults_modelperformances}. In Figure \ref{fig:dt_tuning} we present the results of the decision tree's tuning process. By plotting the MSE obtained on the test and training set, we visualize that the model overfits for a depth higher than five. In Figure \ref{fig:rf_tuning} we plot the MSE obtained by the random forest on the test set for the considered ensemble sizes and depths. In line with the results observed for the decision tree, we see that the random forest is also superior for small depths. We achieve the lowest MSE for a depth of eight and an ensemble size of 50. 
\input{Tables/Space_CNN}
\input{Tables/Space_MLP}
\input{Tables/Space_DT_RF.tex}

%% file: Tables/cnn_architecture.tex
\begin{table}[ht]
\centering
\begin{adjustbox}{width=.97\textwidth}
\begin{tabular}{l | c | c | c | l | l}

Layer&\#Layers&Output Shape&  \#Parameters & Connected To\\\hline

InputLayer3D         &$1$         & $q \times p \times L \times 1$                         & 0 &- \\

$\text{Conv3D}_1$    &$1$         & $q \times p \times L \times M_{[\text{Conv3D}_1]}$      & eq. (\ref{eq:nrparam3dconv})& InputLayer3D\\

$\text{Conv3D}_2$    &$1$         & $q \times p \times L \times M_{[\text{Conv3D}_2]}$      & eq. (\ref{eq:nrparam3dconv})& $\text{Conv3D}_1$\\

InputLayer3DTrans    &$|{X^{\text{1D}}_{i}}^{\text{T}}|$ & $1 \times 1 \times L \times 1$                          &0&-\\

Conv3DTrans          &$|{X^{\text{1D}}_{i}}^{\text{T}}|$ & $q \times p \times L \times M_{[\text{Conv3DTrans}]}$   & eq. (\ref{eq:nrparam3dconv})& InputLayer3DTrans\\

Concatenation3D      &$1$         & $q \times p \times L \times (M_{[\text{Conv3D}_2]} + M_{[\text{Conv3DTrans}]} |{{X^{\text{1D}}_{i}}}^{\text{T}}|)$ &   0   & $\text{Conv3D}_2$, Conv3DTrans\\

$\text{Conv3D}_3$    &$1$         & $q \times p \times 1 \times M_{[\text{Conv3D}_3]}$                & eq. (\ref{eq:nrparam3dconv}) & Concatenation3D\\

$\text{Reshape}_1$   &$1$         & $q \times p \times M_{[\text{Conv3D}_3]}$                         &0&$\text{Conv3D}_3$\\

InputLayer2D         &$|X^{\text{2D}}_{i}|$ & $q \times p \times 1$                                             &0&-\\

Concatenation2D      &$1$         & $q \times p \times (M_{[\text{Conv3D}_3]} + |X^{\text{2D}}_{i}|)$             &0&$\text{Reshape}_1$, InputLayer2D\\

LocallyConnected2D   &$1 $        & $q \times p \times 1$                                             & eq. (\ref{eq:nrparamlocconn})&-\\

$\text{Reshape}_2$   &$1$         & $qp$                                                              &0&Concatenation2D\\

InputLayer1D         &$|{X^{\text{1D}}_{i}}^{\Bar{\text{T}}}|+|X^{\text{S}}_{i}|$ & $l_j \ \forall x_{{i}_j} \in {X^{\text{1D}}_{i}}^{\Bar{\text{T}}}, 1 \ \forall x_{{i}_j} \in X^{\text{S}}_{i}$                                         &0&-\\

Concatenation1D        &$1$       & $qp + \sum_{x_{{i}_j} \in {X^{\text{1D}}_{i}}^{\Bar{\text{T}}}} l_j + |X^{\text{S}}_{i}|$                                     &0&$\text{Reshape}_2$, InputLayer1D\\

$\text{DenseLayer}_1$       &$1$       & $N^{[\text{DenseLayer}_1]}$                                            &eq. (\ref{eq:nrparamdense})&Concatenation1D\\

$\text{DenseLayer}_2$       &$1$       & $N^{[\text{DenseLayer}_2]}=qp$                                         &eq. (\ref{eq:nrparamdense})&$\text{DenseLayer}_1$\\

$\text{Reshape}_3$     &$1$       & $q \times p \times 1$                                                &$0$&$\text{DenseLayer}_2$\\

\end{tabular}
\end{adjustbox}
\caption{{CNN Architecture}}
\label{table:cnn_architecture}
\end{table}

%% file: Tables/Space_CNN.tex
\begin{table}[h!]
\centering
\begin{adjustbox}{width=.8\textwidth}
\begin{tabular}{l | l | l}

Hyperparamaters ($\theta$)&Space ($\Theta$)&Set ($k$)\\\hline
Number of filters $Conv3D_1$, $Conv3D_2$, $Conv3D_3$ & \{1,2,...,128\}&k=1\\
Number of filters $Conv3DTrans$ & \{1,2,...,128\}&k=1\\
Number of units $DenseLayer_1$ & \{1,2,...,240\}&k=1\\
Kernel size $Conv3D_1$, $Conv3D_2$ & \{2,3\}&k=1\\

\hline

Activation function $Conv3D_1$, $Conv3D_2$, $Conv3D_3$ &\{relu,elu,leakyrelu\}&k=2\\
Activation function $Conv3DTrans$&\{relu,elu,leakyrelu\}&k=2\\
Activation function $DenseLayer_1$, $DenseLayer_2$ &\{relu,elu,leakyrelu\}&k=2\\
Activation function $LocallyConnected2D$ &\{relu,elu,leakyrelu\}&k=2\\

\hline

Batch size&\{32,64,128\}&k=3\\
Dropout $DenseLayer_1$ & [0,0.5]&k=3\\
L2 kernel regularizer $Conv3D_1$, $Conv3D_2$, $Conv3D_3$ & [0,0.5]&k=3\\
Learning rate & [1E-06;0.1]&k=3\\
Optimizer & \{Adam, SGD\}&k=3\\

\end{tabular}
\end{adjustbox}
\caption{CNN Hyperparameter Domains and Set Assignments for the hierarchical BO}
\label{table:CNN_Search_Space}
\end{table}

%% file: Tables/Space_MLP.tex
\begin{table}[h!]
\centering
\begin{adjustbox}{width=.8\textwidth}
\begin{tabular}{l | l | l}

Hyperparamaters ($\theta$)&Space ($\Theta$)&Set ($k$)\\\hline

Number of hidden layers & \{1,2,3,4\}&k=1\\
Number of units in hidden layers & \{1,2,...,100\}&k=1\\

\hline

Activations in hidden layers \& output layer &\{relu,elu,leakyrelu\}&k=2\\

\hline

Dropout in hidden layers & [0,0.5]&k=3\\
Batch size&\{32,64,128,256\}&k=3\\
Learning rate & [1E-06;1E-04]&k=3\\
Optimizer & \{Adam, SGD\}&k=3\\

\end{tabular}
\end{adjustbox}
\caption{MLP Hyperparameter Domains and Set Assignments for the hierarchical BO}
\label{table:MLP_Search_Space}
\end{table}

%% file: Tables/Space_DT_RF.tex
\begin{table}[h!]

\centering
\begin{adjustbox}{width=.6\textwidth}
\begin{tabular}{l | l | l }
Model & Hyperparameters ($\theta$)&Space ($\Theta$)\\\hline

Decision Tree & Max. Depth&\{2,4,...,100\}\\\hline

\multirow{2}{*}{Random Forest}& Max. Depth&\{2,4,...,50\}\\
&Number of Trees&\{50,100,...,1000\}

\end{tabular}
\end{adjustbox}
\caption{Grid Search: Random Forests and Decision Trees
}
\label{table:GS_Trees}
\end{table}